\documentclass[twoside]{article}

\usepackage[accepted]{aistats2025}
\usepackage{hyperref}
\usepackage{url}
\usepackage[table]{xcolor}         
\usepackage{multirow}      
\usepackage{enumitem}       
\usepackage{graphicx}  
\usepackage{tabularx}
\usepackage{subcaption} 
\usepackage{xcolor}  
\usepackage{amsmath}

\usepackage{multicol}
\usepackage{multirow}
\usepackage{amsthm}
\usepackage[capitalise]{cleveref}
\usepackage{subcaption}
\usepackage{wrapfig,lipsum,booktabs}
\usepackage{comment}
\usepackage{placeins} 
\usepackage{float}
\usepackage{array} 

\newcommand{\fractaldim}{\ensuremath{D_f}}
\newcommand{\scalefreev}{\ensuremath{\alpha_v}}
\newcommand{\scalefreec}{\ensuremath{\alpha_c}}
\newcommand{\modularity}{\ensuremath{Q}}

\newcommand{\backbone}{\ensuremath{B_b}}
\newcommand{\treewidth}{\ensuremath{T_w}}

\newcommand{\centralityb}{\ensuremath{B_e}}
\newcommand{\Proximaty}{\ensuremath{P_r}}
\newcommand{\Entropy}{\ensuremath{H}}

\begin{document}

\twocolumn[

\aistatstitle{Structure based SAT dataset for analysing GNN generalisation}

\aistatsauthor{ Yi Fu \And Anthony Tompkins \And Yang Song \And Maurice Pagnucco }

\aistatsaddress{ School of Computer Science and Engineering, University of New South Wales, Australia} ]

\begin{abstract}
Satisfiability (SAT) solvers based on techniques such as conflict driven clause learning (CDCL) have produced excellent performance on both synthetic and real world industrial problems. 
While these CDCL solvers only operate on a per-problem basis, graph neural network (GNN) based solvers bring new benefits to the field by allowing practitioners to exploit knowledge gained from solved problems to expedite solving of new SAT problems.  However, one specific area that is often studied in the context of CDCL solvers, but largely overlooked in GNN solvers, is the relationship between graph theoretic measure of \textit{structure} in SAT problems and the \textit{generalisation} ability of GNN solvers. To bridge the gap between structural graph properties (e.g., modularity, self-similarity) and the generalisability (or lack thereof) of GNN based SAT solvers, we present \textbf{StructureSAT}: a curated dataset, along with code to further generate novel examples, containing a diverse set of SAT problems from well known problem domains. Furthermore, we utilise a novel splitting method that focuses on deconstructing the families into more detailed hierarchies based on their structural properties. With the new dataset, we aim to help explain problematic generalisation in existing GNN SAT solvers
by exploiting knowledge of structural graph properties. We conclude with multiple future directions that can help researchers in GNN based SAT solving develop more effective and generalisable SAT solvers. 
\end{abstract}

\section{Introduction}

The satisfiability (SAT) \cite{biereHandbookSatisfiability2009} problem is a hallmark of computer science research with remarkable real-world utility, especially in solving combinatorial optimisation problems. SAT forms the basis of the study of computational complexity especially around the complexity class of NP problems. 
Moreover, as a theoretical tool, it facilitate research into the nature of computation and solving difficult computational problems. 
On the practical side, SAT has been applied to many interesting real world problems such as logistics planning \cite{kautzUnifyingSATbasedGraphbased1999}, product configuration \cite{sinzFormalMethodsValidation2003}, and software verification \cite{ivancicEfficientSATbasedBounded2008}, 
retaining its high relevance today.

Advances over the last decades in SAT solving have appeared to converge on conflict driven clause learning (CDCL) methods \cite{marques-silvaChapterConflictDrivenClause2021} for the best general problem solving performance.
While it is widely believed that CDCL solvers are performant towards various aspects of problems~\cite{alyahyaStructureBooleanSatisfiability2023}, such solvers almost exclusively operate on a per-problem basis. That is to say they do not explicitly reuse knowledge from different problems.
Alternatively, graph neural networks (GNNs) have emerged as a complementary approach to representing and solving SAT problems by incorporating the benefits of deep learning \cite{guo2023machine}.
Using an optimisation based approach, as opposed to a pure search like algorithm in CDCL, GNN-based solvers have the potential to adapt useful information from training problems to accelerate solving unseen novel problems.

However, SAT problems largely reside in NP-hard problems. With machine learning methods such as GNNs, prior works have demonstrated provably negative results on challenges that are NP-hard \cite{yehudaItNotWhat2020}. 
Although deep neural networks have the ability to ingest large datasets ~\cite{krizhevsky2017imagenet,simonyan2015deep}, there lacks a dataset that facilitates sufficiently large and unbiased training for SAT problems, and existing SAT datasets with higher difficulties generally have a limited number of problems. Indeed, currently the largest benchmarks SATLIB ~\cite{hoosSATLIBOnlineResource} and SATCOMP \cite{InternationalSATCompetition} contain less than 10k industrial problems.
While we can generate synthetic datasets to train GNNs~\cite{selsam2018learning}, the resulting models would struggle to generalise to more diverse and challenging, or real-world problems~\cite{liG4SATBenchBenchmarkingAdvancing2023a}, limiting the application of GNN-based solvers.

Moreover, currently, the generalisability of GNN solvers has been significantly overlooked, especially in terms of the relationship between generalisation and graph structures of SAT problems.
More specifically, SAT problems in prior work are typically generated in a random manner without delving into what makes instances difficult or useful for training and testing, potentially limiting the models from generalising to more challenging datasets, especially to industrial instances.
For example, the largest SAT dataset on GNN -- G4SATBench \cite{liG4SATBenchBenchmarkingAdvancing2023a}, which is constructed with 7 generators, only considers generalisation regarding the numbers of variables as a measure of performance vs.\ complexity. The impact of the \textit{structural properties} of a dataset is however ignored and remains a less explored area. In particular, problem difficulty can be strongly influenced by the intrinsic \textit{graph structure} of each problems, as shown experimentally using traditional solver \cite{alyahyaStructureBooleanSatisfiability2023}. It is thus intuitive to hypothesise that such structural properties studied in SAT would influence GNN's performance, especially their ability to generalise. However, for most GNN solvers, only the ability to generalise to larger, in-distribution problems from the same domain is discussed, whereas the structural properties across domains are not investigated.

To bridge the gap between structural measures and generalisability of GNNs on SAT, we propose StructureSAT, a large-scale dataset containing diverse problem domains and structural measures. In this work, we are not interested in the set of all possible SAT problems, but rather existing, well studied, SAT problem domains which we aim to analyse through the lens of graph theoretic structure. 
Thus, we focus on easy-to-generate synthetic training datasets for investigating GNN's generalisability.
StructureSAT contains 11 SAT domains from 4 high level categories: random, crafted, pseudo-industrial and industrial, within which we study 9 structural properties that have proven to be influential to traditional SAT solvers, including both conjunctive normal form (CNF) based and graph based properties~\cite{alyahyaStructureBooleanSatisfiability2023}. 
With traditional SAT solvers, typically certain structure values are controlled for each domain, and CDCL solving time are recorded as a metric of solver performance \cite{giraldez2016generating}, so that the effects of different structural properties can be analysed using different domains. However, it is difficult and expensive to follow this approach in GNN-SAT solving given the vast amount of combination of choices in training and testing domains, structure values, models used and evaluation metrics. To make this task manageable, in this work, we generate different groups of training and testing sets based on the structure values, and evaluate the GNN generalisability for both in-domain and out-domain distributions. 
Specifically, we carefully deconstruct each problem domains based on their structural properties, and split each domain into multiple subsets based on the range of values for each of the properties. Based on these subsets, we analyse the relationship between the SAT problem structures and generalisation ability of GNN solvers with three GNN models. This analysis is conducted on a diverse set of in-domain (training/validation and testing sets from the same domain but with different structural ranges) and out-domain problems (training and testing sets from different domains) to comprehensively evaluate different scenarios for generalisation. 
Our results demonstrate that GNN solvers provide higher generalisability for certain structural properties but significantly poor performance for others, thus highlighting potential future directions on improving the generalisation performance of current GNN-based solvers. The main contributions of this paper are as follows:

\begin{itemize}[noitemsep, topsep=0pt, leftmargin=*]
\item We present StructureSAT, a large-scale dataset for investigating generalisation capability of GNN-based SAT solvers. 
\item We provide a comprehensive analysis on the properties of StructureSAT and problem difficulty, demonstrating that different graph structures have different effects on the generalisability of GNN solvers.
\item We conduct extensive experiments with GNNs, focusing on how structural properties of each problem domain affect different  generalisation tasks with both in-domain and out-of-domain SAT problems. Our results indicate that significant future work is required to enhance the generalisation performance of GNN-based SAT solvers.
\end{itemize}

\section{Related work}
\textbf{SAT dataset.}
SAT problems can be classified into different categories, mainly as random, crafted, and industrial \cite{alyahya2022structure}, while most come from either synthetic generators or existing datasets. Most generators such as CNFgen \cite{lauriaCNFgenGeneratorCrafted2017}, focus on random problems like random \textit{k}-SAT or combinatorial problems. The largest established datasets are SATLIB \cite{hoosSATLIBOnlineResource} and SAT Competitions (SATCOMP) \cite{InternationalSATCompetition}, while only SATCOMP contains industrial problems and is limited in size. To overcome this constraint, new generators have been proposed to generate pseudo-industrial instances, including hand-crafted generators Community Attachment \cite{giraldez-cruModularityBasedRandomSAT2015} and Popularity-Similarity  \cite{giraldez-cruLocalityRandomSAT2017}, or graph-generative models generators \cite{youG2SATLearningGenerate2019,li2024distribution,li2023hardsatgen,chen2023matching,garzon2022performance}. For GNN based SAT solving, \cite{selsam2018learning} proposed random generator SR(\textit{n}) and \cite{cameronPredictingPropositionalSatisfiability2020} generated uniform-random 3-SAT instances as datasets. G4SATBench \cite{liG4SATBenchBenchmarkingAdvancing2023a} produced a dataset including 7 types of synthetic generators with varying variable sizes.

\textbf{SAT structural properties.}
As traditional SAT solvers perform differently over random, crafted, and industrial instances, it is believed that different SAT domains have distinct underlying properties \cite{alyahyaStructureBooleanSatisfiability2023}, including problem-based properties and solver-based properties. Results in this filed have been wildy used in improving traditional solvers \cite{audemard2009predicting} and portfolio based solvers \cite{sonobe2016cbpenelope2016}. 
Problem-based properties are further divided to CNF-based, including phase-transition \cite{cheeseman1991really}, backdoor \cite{kilby2005backbones} and  backbones \cite{kilby2005backbones}, and graph-based, including scale-free \cite{ansoteguiStructureIndustrialSAT2009}, self-similar \cite{ansoteguiFractalDimensionSAT2014}, centrality \cite{katsirelos2012eigenvector}, small-worlds \cite{walsh1999search} and community structure \cite{ansoteguiCommunityStructureIndustrial2019}. Solver-based properties are related to SAT solvers such as: mergeability and resolvability \cite{zulkoski2018effect}. These measures are either proved mathematically, or analyzed through solver-related parameters such as solving time. However, all the works are experimented with traditional SAT solvers, and most work only focus on single property measure \cite{li2021hierarchical, zulkoski2018effect}. In this work, we select several related properties, and calculate their values on each of our selected domains. We also split our dataset to different values based on each structure, 
and analyze the important or hard features for GNN to capture. 

\textbf{GNN for SAT solving.} GNN has been applied to SAT solving mainly as problem solvers, including standalone solvers, which are networks trained to classify problem satisfiability themselves and mainly encodes the formula as LCG graphs \cite{selsam2018learning,zhangBetterGeneralizationNeural2022,changPredictingPropositionalSatisfiability2022,shi2023satformer,cameronPredictingPropositionalSatisfiability2020,hartfordDeepModelsInteractions2018,duanAugmentCareContrastive2022,ozolinsGoalAwareNeuralSAT2022}, and hybrid solvers, which treat GNN as a guidance to traditional solvers by replacing their heuristics with network predictions. These solvers focuses on specific tasks and corresponding problems such as UNSAT core \cite{selsam2019guiding}, glue clauses \cite{hanEnhancingSATSolvers2020} or backbone variables \cite{wangNeuroBackImprovingCDCL2023} for CDCL solvers and initial assignment \cite{zhangNLocalSATBoostingLocal2020, liNSNetGeneralNeural2022} for SLS solvers. Although hybrid solvers generally achieve better results than standalone solvers, they acts as modifications of traditional solvers instead of discussing solvability of GNNs,  which is out of scope of our work. 

\section{Dataset description}
\label{section:dataset}

StructureSAT consists of multiple problem domains, for each of which we control the corresponding attribute values.
In this section, we start by introducing the problem definition of satisfiability (SAT). Then, we present the graph structure properties measured in the dataset. Next, we introduce the types of data and generators used for raw data generation. Last but not least, we detail the construction process, structure-based splitting methods, and generalization tasks.

\subsection{SAT preliminaries}
In propositional logic, the SAT problem is the problem of finding an assignment of truth values (true or false) to propositional variables that make a Boolean formula satisfiable (i.e., true).  Boolean formulae are typically expressed in conjunctive normal form (CNF). We denote the set of propositional variables by $V$. A literal $l$ is either a variable $v \in V$ or its negation $\neg{v}$ (or $\overline{v}$). A clause $c$ is a disjunction of literals ($l_1 \lor l_2 \lor.. l_n$), where $\lor$ denotes propositional logic connective ``or". A formula $f$ is a conjunction of clauses $(c_1 \land c_2 \land... c_n)$, where $\land$ denotes propositional logic connective ``and". Generally speaking classical SAT solvers can be divided into \emph{complete solvers} and \emph{incomplete solvers}. A complete solver is able to prove unsatisfiability or find a satisfying assignment if they exist for a problem. Most complete solvers are based on the Davis–Putnam–Logemann–Loveland (DPLL) algorithm \cite{davisComputingProcedureQuantification1960,davisMachineProgramTheoremproving1962}, a backtracking search algorithm. Two popular types of solvers built from the DPLL solver are the Conflict-Driven Clause Learning (CDCL) solvers  \cite{marques-silvaChapterConflictDrivenClause2021}, which learn and add new conflict clauses to the original formula, and look-ahead solvers  \cite{heuleCubeConquerGuiding2012}. 
In contrast, an incomplete solver cannot prove unsatisfiability such as the Moser-Tardos (MT) solver \cite{catarataMoserTardosResampleAlgorithm2017} and WalkSAT solver \cite{selman1994NoiseStrategiesImprovinga} that are derived from the stochastic local search (SLS) algorithm.

SAT formulas have been expressed using various graphs 
\cite{alyahyaStructureBooleanSatisfiability2023}. Traditional SAT community usually encodes SAT formulas as undirected weighted graph including \textit{variable incidence graph} (VIG) and \textit{clause-variable incidence graph} (CVIG) \cite{ansoteguiCommunityStructureSAT2012}. VIG is a graph with variables as nodes and two literals are connected with edges if they occur in the same clauses. CVIG is a bipartite graph having variables and clauses as vertices. They are connected if a variable occurs inside a clause. Another commonly used set of graphs are unweighted graphs including VIG, \emph{variable clause graphs} (VCG), \emph{literal incidence graphs} (LIG), and \emph{literal clause graphs} (LCG). LIG extends VIG through extra edges between literals and their negations. VCG and LCG are bipartite graphs similar to CVIG, but without weights. LCG also has an extra edge between literals and their complements. Among all the graphs, LCG has been used most in GNN based SAT solving as others either lose important information of clause (LIG and VIG), or information of polarity (VCG). As prior research \cite{liG4SATBenchBenchmarkingAdvancing2023a} has shown that the particular bipartite graph construction (LCG or VCG) does not have much different effect on model accuracy while non-bipartite graphs (LIG or VIG) could not represent SAT well for GNN training, in this work we will be focusing on LCG only.

\subsection{SAT structure properties}
\label{tab:structures}
In this work, we choose several structural properties from prior studies based on two criteria: public availability of their code and algorithms, and their proven impact on traditional CDCL solvers. We then apply these properties as a splitting method for training and testing in the dataset. StructureSAT classify structural properties of SAT problems into two classes based on the embedding methods: CNF based properties and graph based properties \cite{alyahya2022structure}. 

\begin{figure*}[t]

    \centering
    \includegraphics[width=0.87\textwidth]{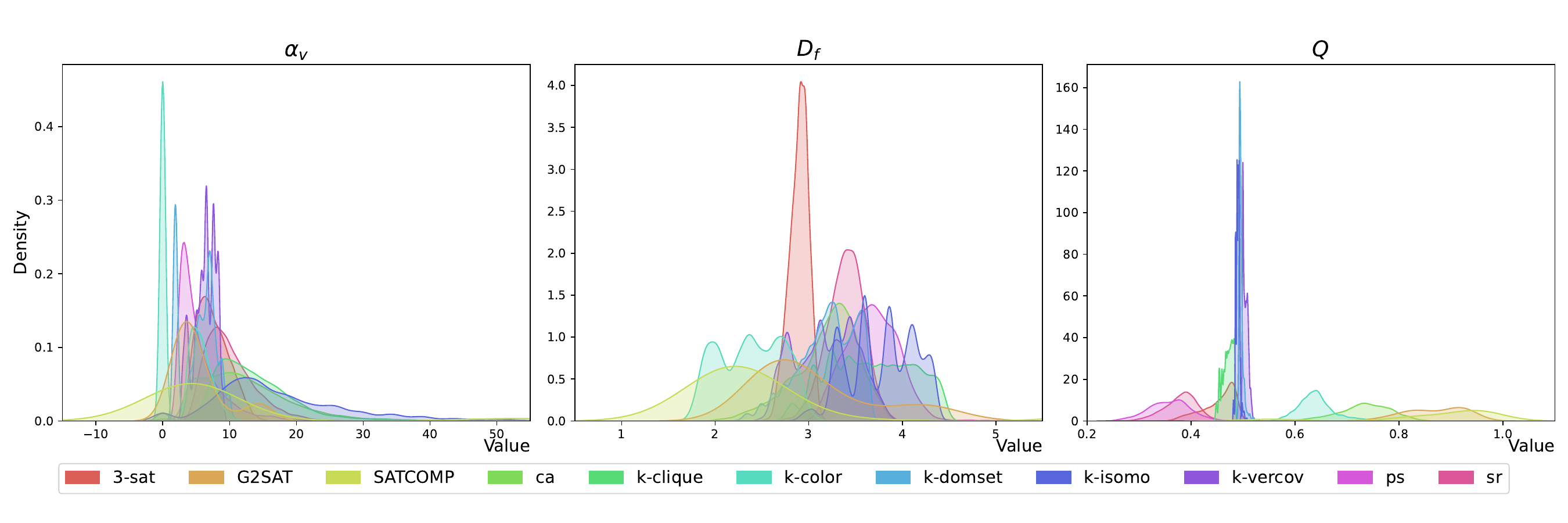}
    \vspace{-5pt}
    \caption{Distributions of various structural graph properties from different problem domains. Left: scale-free measure by variable $\scalefreev$, Middle: fractal dimension $\fractaldim$, Right: modularity $\modularity$.}
    \label{fig:example}
    \vspace{-10pt}
\end{figure*}

\textbf{CNF based properties.} Given a problem $P$ with set of variables $V$, the CNF based properties are related to the CNF encoding of a problem. 
\begin{itemize}[leftmargin=*]

      \item \textbf{Phase transition} \cite{cheeseman1991really} is a phenomenon measured by clause to variable ratio ($n_c/n_v$), where $n_c$ and $n_v$ represents number of clauses and variables respectively. It is evident that an easy-hard-easy pattern occurs for random \textit{k}-SAT, where transition for random 3-SAT is at $c = 4.258*n + 58.26 * n ^ {-1/2} $ \cite{crawford1996experimental}. Although the phenomenon has been observed in both random \textit{k}-SAT and peudo-industrial instances, for simplicity we will only be focusing it on random 3-SAT in this dataset. 

      \item \textbf{Backbone} \cite{kilby2005backbones} refers to the set of literals of a problem which are true in all satisfying assignments. The value of each literals in the set is fixed in all assignments of the problem. StructureSAT calculates the size of the backbones($\backbone$) of satisfiable instances only.
    
\end{itemize}

\textbf{Graph based properties} are structural properties from embeddings of a \textit{LCG} graph $G$. 
Given a problem with set of variables $V$ and set of clauses $C$, the LCG graph $G$ with vertex $X$ and weight $w$ is defined as $G = (X, {x|x\in C \land V}, w)$ and the weight is either $1/|X|$ if variable is in clause, $1/|V|$ for weights between variables and their negations, or 0 if nodes are not connected. Each node $x$ in graph has degree $deg_x$.
\begin{itemize}[leftmargin=*]
 \item \textbf{Self-similary} \cite{ansoteguiFractalDimensionSAT2014} is measured by fractal dimension({\ensuremath{D_f}}). $G$ is self-similar if the minimum number of boxes of size $s$ required to cover $G$ decreases polynomially for some {\ensuremath{D_f}}.
    \item \textbf{Scale-free} \cite{ansoteguiStructureIndustrialSAT2009} can be measured by frequency of variables ({\ensuremath{\alpha_v}}) or clauses size ({\ensuremath{\alpha_c}}). A graph is scale-free if the arity of nodes is characterized by a random variable $N$ that follows a power-law distribution.
    \item \textbf{Treewidth} ($\treewidth$) \cite{mateescu2011treewidth} measures the tree-likeness of graphs. It is the minimum width over all possible tree decomposition of $G$. In this work we calculate treewidth using $treewidth-min-degree$ function from NetworkX \cite{hagberg2020networkx}, which calculates using the Minimum Degree heuristic.
    \item \textbf{Centrality} specifies how important a node is within a graph. Following \cite{freeman1977set}, we calculate the betweeness centrality $BE$ using NetworkX \cite{hagberg2020networkx}.
    \item \textbf{Community structure} \cite{ansoteguiCommunityStructureIndustrial2019} is measured by modularity($Q$). The modularity $Q$ of $G$ is computed with respect to a given partition $C$ of the same graph and is a measure of the fraction of within-community edges in relation to \textit{another} random graph that has an equal number of vertices and degree. A graph's modularity score corresponds to the maximal modularity of any possible partition in C: $Q(G) = \max\{Q(G,C) | C\}$. While computing the exact value of $Q$ is NP-hard, we would approximate lower-bounds to $Q$.

     \item \textbf{Small-world} \cite{walsh1999search} measures the extent of graph topology. It is approximated using Proximity Ratio ($\Proximaty$). $G$ is considered small-world if $\Proximaty \geq 1$. In this work we use the sigma function from NetworkX \cite{hagberg2020networkx} for calculation, which is only an approximation of the $\Proximaty$ value due to the need for random graph generation.
     \item \textbf{Entropy} \cite{zhang2021structural} measures the uncertainty of random systems. Given volume of the graph $vol_G$, we measure the One-dimensional Entropy ($\Entropy$) following \cite{zhang2021structural}.

\end{itemize}

\subsection{StructureSAT composition}
\subsubsection{Dataset generation}
\label{sec:rawd}

We use various existing SAT generators and datasets to produce original SAT formulas in 11 domains. To show the variety of structural differences,  
we show the distributions of various structural properties  in \cref{fig:example} from calculating the statistical values on raw data pairs with a 60 seconds timeout. The distributions are plotted on 10k problem pairs for each selected domains, with each pair containing 50\% satisfiable and 50\% unsatisfiable  problems. 
  For industrial SAT problems and graph generative model generators, less than 100 pairs are selected within the time limit. Next,we detail the provided domains in StructureSAT, including important parameters used for generation. Note that we create 5 domains from combinational problems, and other approaches produce one domain each.

\textbf{Random-3-SAT}: Uniform random 3-SAT ($r$-$3$SAT) is a special case for Random \textit{k}-SAT where each clause contains exactly 3 literals. The problems are generated using CNFgen \cite{lauriaCNFgenGeneratorCrafted2017} with $10-40$ number of variables at the phase transition point.

\textbf{SR(\textit{n})}: SR(\textit{n}) \cite{selsam2018learning} is a special expression of random $k$-SAT that consists of a balanced dataset with pairs, with $k$ as the maximum number of variables in the problem. Each pair contains one satisfiable problem and one unsatisfiable problem, differing by 1 single literal in one clause. In this work we use problem with $10-40$ number of variables as main training data, represented as SR(10-40).

\textbf{Combinatorial problems} (5 domains): For most combinatorial problems, the goal is to find some combination of elements in a solution space while respecting defined constraints. A valid solution would correspond to a satisfying assignment in the SAT encoding of the problem \cite{biere2009handbook}. In StructureSAT, we generate 5 combinatorial problems using CNFgen \cite{lauriaCNFgenGeneratorCrafted2017}. Each problem is related to solving constraints over a graph. Thus, we generate random graphs using the Erdős–Rényi model \cite{newman2018networks} with edge probability $\binom{v}{k}^{-1/\binom{v}{2}}$, where $v$ representing number of vertex. The selected parameters and problems include $k$-coloring, $ 3\leq k \leq 4$, $k$-dominating-set, $ 2\leq k \leq 3$, $k$-clique-detection, $ 3\leq k \leq 4$, $k$-vertex-cover, $ 3\leq k \leq 5$, and automorphism in graph \textit{G}. 

\textbf{CA}: Community Attachment (CA) is a seminal work generated from \cite{giraldez-cruModularityBasedRandomSAT2015} to mimic the community structure of industrial problems, i.e., value of \textit{modularity} $Q$ on a SAT instance's VIG graph. In StuctureSAT, we select 0.7-0.9 as the value of $Q$. The generator uniformly and randomly selects 4 to 5 literals inside the same community with probability $P = Q + 1/co$, where $co$ is the number of communities, and 4 to 5 literals in distinct community with probability $1 - P$. These process are done iteratively to form a formula.

\textbf{PS}: Popularity-Similarity (PS) \cite{giraldez-cruLocalityRandomSAT2017} is a generation algorithm developed on the idea of \textit{locality}, which is a measure of both community structure and scale-free structure on a SAT instance's VCG graph. PS randomly samples variable $v$ in clause $c$ with the probability $P = 1/(1+n_v^\beta * n_c^{\beta_1} * \theta_{n_v*n_c}/R)^T$, where $\beta$ is power-law distribution, $\theta$ is random angle assigned to $v$ and $c$, $T$ is temperature between 0.75 and 1.5, and $R$ is an approximate normalisation constant.

\textbf{SATCOMP}: StructureSAT uses industrial instances 
from SAT competition (SATCOMP) 2007. 

\textbf{G2SAT}: To overcome the limiting number of instances in industrial problem, graph generative models have been used as problem generator. In this work we mainly use \textbf{G2SAT} \cite{youG2SATLearningGenerate2019} as generator. Following previous work, we select problems from the industrial dataset, standadise with SatElite preprocessor\cite{een2005effective}, and generate similar instances using GraphSAGE \cite{hamilton2017inductive} and a two-phase generation process. Both industrial and G2SAT datasets are used mainly as testing set in the dataset.

\subsection{Augmented problems}
This section describes augmented dataset setup that might affect the generalisation ability of GNN. CDCL-based SAT solvers produce conflict clauses during searching and add learned clauses, which are reverse of conflict clauses, to the original problems. Prior work \cite{liG4SATBenchBenchmarkingAdvancing2023a} has shown that training on augmented SAT problems with the learned clauses added could lead to better accuracy on augmented SAT domains than training on raw problems. Since this might be caused by the destroying of structure during the addition of learned clauses \cite{ansoteguiCommunityStructureIndustrial2019}, we are interested in an empirical analysis on the structural difference and their effect on out-of-distribution generalisation. Thus, we gather learned clauses from DRAT-trim proofs \cite{wetzler2014drat} after running problems with CadiCal solver \cite{fleury2020cadical}, then augment our dataset with the collected learned clauses.

\subsection{Structure-aware splitting and generalisation selection}
\label{sec:structure-aware-splitting}
We propose a novel splitting method for 
StructureSAT
and focus on 3 types of generalisation tasks. 
To do this, after generating the raw datasets from each domain, instead of randomly splitting them to train/valid/test sets, we divide each raw dataset based on specific structural values. Specifically, for each domain, we select the average value $Z$ from every calculated graph-based properties, and split each domain into a low-value subset and a high-value subset, producing a total of 16 subsets (8 structures $*$ 2 ranges) 
per domain. Each subsets contains 80k training data and 10k validation data.\footnote{Our dataset and codebase are available \href{https://drive.google.com/drive/folders/1ZrhRlRqQUTrYExVzVbXaocjvNIi2Os25?usp=sharing.}{here}.\\}

The three types of generalisation tasks are represented using 3 aspects of test problems: in-domain larger problems, in-domain problems with different properties, and out-of-distribution problems. Specifically, for in-domain larger problems, we follow the generation process described in \cref{sec:rawd} and randomly generate larger problem sets $D_{t1}$ using all the synthetic generators. For example, for training with SR problems, the training and validation sets contain problems with 10-40 number of variables, while the testing sets contain problems with 40-100 and 100-200 number of variables, each with 10k pairs.

To test generalisation on in-domain problems with different structures, we are interested in CNF-based properties. Thus we focus on random 3-SAT problems with different $c/v$ ratios and different backbone values on all domains. For analysing the $c/v$ ratio on random 3-SAT, the training and validation sets are problems with 10-40 number of variables, and a $c/v$ ratio of roughly 4.7. Our testing set contain random 3-SAT problems with same number of variables, but different number of clauses. Specifically the test set $D_{t2}$ has a $c/v$ ratio of 3.5, 4, 5.5 and 6, with each ratio having 10k pairs of problems. For backbone on all domains, the test sets comprise solely satisfiable problems, categorized according to the average backbone size following the implementation of \cite{biere2023cadiback}. Additionally, we also include existing dataset - random 3-SAT problems with controlled number of variables (100) and clauses (403) but different backbone sizes sourced from SATLIB \cite{hoosSATLIBOnlineResource}.

For out-of-distribution problems, we randomly generate 10k problem testing data from each synthetic generators without property split and form $D_{t3}$, thus holding the structural range as in \cref{fig:example}. We combine problems from every domain, excluding the training domain, to form the out-of-domain testing sets $D_{t4}$. Small-size industrial and application problems from SATCOMP and generated G2SAT problems are also selected as larger industrial testing set in $D_{t3}$ . 

With the novel splitting and generation methods introduced in StructureSAT, we produces different training, validation and testing sets. These sets comprise: (1) problems with certain attribute fixed at a value, (2) problems with certain attribute fixed within a range, and (3) randomly generated problems without attribute-based splitting. These sets are then used to analyse different generalisation tasks.

\section{Experiment}
\label{tab:experiment}
In this section, we experimentally evaluate the generalisation ability of GNN-based SAT solvers with StructureSAT. We are interested in how is the generalisability of GNN affected by the structures of both training and testing instances. Specifically, we investigate the following questions:  
Q1: How does the structure influence generalisation to in-domain larger size problems;
Q2: Could GNNs generalise to the same domain but for problems with different structure;
Q3: How much does structure influence out of distribution generalisation;
Q4: Does training on augmented problems influence out of distribution generalisation?

\begin{table}[t]
        \centering
        \normalsize
        \caption{{NeuroSAT testing results. 
        For each metric, we train NeuroSAT on the small and big split of the corresponding SR(10-40) training sets. 
        We then test the trained model on the testing data of 10-40, 40-100, and 100-200 variables for each metric. 
        Better training split performance in bold.}}
        \vspace{2.7pt}
        \resizebox{0.48\textwidth}{!}{
        \begin{footnotesize}
        \begin{tabular}{>{\centering\arraybackslash}p{1.2cm} >
        {\centering\arraybackslash}p{2.4cm} >
        {\centering\arraybackslash}p{0.75cm} >
        {\centering\arraybackslash}p{0.85cm} >{\centering\arraybackslash}p{0.85cm} >{\centering\arraybackslash}p{0.85cm} >{\centering\arraybackslash}p{0.85cm} >{\centering\arraybackslash}p{0.85cm}}
        \toprule
           \multirow{2}{*}{Train Split} &
           \multirow{2}{*}{Test Set} &
           \multicolumn{6}{c}{Metric}        \\ \cmidrule(lr) {3-8}
        && $\fractaldim$ & $\scalefreev$ & $\scalefreec$ & $\treewidth$& $\modularity$ &  $\Entropy$\\ \midrule
          small & \multirow{2}{*}{10-40} & 94.5 & \textbf{94.4} & \textbf{95.4} & 95.0 & 93.8 & 94.0 \\ big & & \textbf{95.4} & 93.1 & 95.3 & \textbf{95.3} & \textbf{94.8} & \textbf{95.8}\\ \midrule
         small & \multirow{2}{*}{40-100} &  68.5 & \textbf{69.7} & 70.8 & 65.5 & 70.8  &64.8 \\ big & & \textbf{72.4} & 67.1 & \textbf{73.2} & \textbf{72.1} & \textbf{73.3} & \textbf{73.8}\\ \midrule
         small & \multirow{2}{*}{100-200} & \centering \textbf{57.0} & \textbf{56.8} & 57.7 & 54.7  &\textbf{57.9} &53.8 \\  \centering big & &  57.0 & 54.5 & \textbf{59.0} & \textbf{56.5} & 57.8  &\textbf{58.0}\\ \midrule
    \bottomrule
    \end{tabular}
    \end{footnotesize}}
        \vspace{-0.5cm}
        \label{tab:large}
\end{table}

\subsection{Evaluation setup}
With the newly developed dataset StructureSAT, we conduct several experiments on the task of predicting satisfiablity of SAT problems, which is considered a supervised binary graph-classification task. 
We consider different baseline GNN models NeuroSAT \cite{selsam2018learning}, and GCN \cite{kipf2016semi} and GIN \cite{xu2018powerful} 
in our experiments and follow the implementation of G4SATBench \cite{liG4SATBenchBenchmarkingAdvancing2023a} for LCG message passing.
We refer the reader to \Cref{sec:more_exp_details} for more implementation details. 

As prior work has shown that different GNN models 
perform similarly in general 
\cite{liG4SATBenchBenchmarkingAdvancing2023a},
we want to emphasise that the main focus of this paper is not to find the best performing baseline model. Instead, we aim to find the general trend on generalisation performance of GNN-based SAT solvers with regard to various structural measures.

We experiment on each of the generalisation tasks mentioned in \Cref{sec:structure-aware-splitting}. 
Due to limited space, for Q1 and Q3 we only present part of results using SR(10-40) in this section, which is the dataset with the best cross-domain generalisation ability \cite{liG4SATBenchBenchmarkingAdvancing2023a}. We also include some 
Random-3-SAT
experiments 
due to their special phase-transition property. 
All experimental results on other domains and models can be found in \Cref{sec:appendix_result}. In this section, for simplicity, we make SR stand for SR(10-40), R3 stand for Random-3-SAT, KCL stand for $k$-clique, KD stand for $k$-dominating-set, KV stand for $k$-vertex-cover, KCO stand for $k$-coloring, AM stand for automorphism, G2 stand for G2SAT, and IN stand for industrial problems. Note that for simple computation, we select 24 small problems from G2SAT. 12 industrial problems used to train G2SAT are used as testing domain in this section.
\subsection{Generalisation to in-domain problems}
\begin{wraptable}{r}{0.2\textwidth}
\centering
\normalsize
\vspace{-0.3cm}
\caption{R3 results with different ratios.}
    \begin{footnotesize}
    \begin{tabular}{lccc}
    \toprule
    Metric & Accuracy \\
    \midrule
    3.5        & 97.6     \\
    4          & 95.9    \\
    4.7   & 95.5     \\
    5.5        & 98.7     \\
    6          & 99.4    \\
    \bottomrule
\end{tabular}
    \end{footnotesize}
\vspace{-0.1cm}
\label{tab:3sat}
\end{wraptable}
\paragraph{Large size generalisation. }
To answer Q1, we train a NeuroSAT model on a small SAT problem set and test its performance on varying number of variables from testing set $D_{t1}$. \cref{tab:large} shows the testing result of NeuroSAT trained on SR(10-40), which is SR(n) problems with 10-40 number of variables.
In general, larger structural valued problems results in better results on large problems. NeuroSAT are influenced more on certain properties when testing on larger test sets. Specifically, when testing on SR(40-100), $\Entropy$ and $\treewidth$ gets a 9.0\% and a 6.7\% increase when trained on larger splits. Additionally, when comparing between number of variables of test sets, medium sized test set, SR(40-100), has the most difference between large and small split.
We hypothesize this may due to the increase in relevant metric values in larger problems, that makes GNN find more similarity in medium sized properties. Thus, generating problems with property values in the upper range, especially larger $\Entropy$, $\treewidth$ and $\scalefreec$ values as seen in \cref{tab:large} has the potential to enable GNNs solve larger in-domain problems.

\paragraph{Different ratio generalisation. }
\cref{tab:3sat} shows the result of NeuroSAT trained on random 3-SAT and tested on varying $c/v$ ratios from $D_{t2}$, through which we try to answer Q2. Surprisingly, although models are trained with $c/v$ of 4.7, which is the ``hardest" point for random 3-SAT as discussed in \cref{tab:structures}, testing at this point has the lowest accuracy. Furthermore, the further away $c/v$ is from the phase transition point, the better their testing accuracy. This signals that GNNs are not always better at solving in-domain problems with similar properties to the training data. Especially for random 3-SAT problems, GNN would captures the structure of ``simpler-ratioed" R3 problems more easily than the ``harder" ones.

\subsection{Generalisation to other domains}
\label{tab:outofdomain}
 \paragraph{Out-of-distribution generalisation without augmentation}
 To answer Q3, we train different models on the small and big splits of different domains, and test on $D_{t4}$, which contains 8 domains outside of the training domains. Since $D_{t4}$ contains multiple domains, the set contains a wide range of values for each structure. \cref{tab:general} presents the mean and standard deviation of the results across 3 runs trained using SR(10-40) on 3 models. We refer the reader to \Cref{sec:appendix_result} for additional experimental results. in \cref{tab:general}, The accuracy differences differ across the structures. For different models, changing certain structural values in the training set would influence the testing accuracy to different extent. For instance, there is a higher difference between $\fractaldim$ splits on GIN(3.71\%) than on GCN(1.6\%), meaning that $\fractaldim$ is more important in the GIN learning process than GCN. This suggest different models would focus more on capturing certain structures of the dataset. Within the 3 models, $\modularity$ seems to have limited influence on testing accuracy, while $\Entropy$ influences the all models's performance to a greater extent. Moreover, for each model, the structural value influences are different. For GIN, bigger $\scalefreev$ groups achieve better results while smaller $\scalefreec$ groups seems to be better generalised.

To further explore the impact of structure on generalisation, we investigate how similarities in specific structural values between the training and testing datasets affect testing accuracy. We train a GIN model on different groups of SR(10-40) and evaluate its performance on $D_{t3}$. The average results across 3 runs are presented in \cref{tab:general2}. To better understand the structural differences between the training and testing domains, we plotted the average of all structural values and would refer readers to \cref{fulldata} for detailed analysis. 
Overall, the GIN model learns from the input SR structures and generalises more effectively to out-domain problems with similar structural characteristics. For example, the testing set R3 has a smaller average $\fractaldim$ compared to SR, while PS has a smaller $\modularity$ than SR. Models trained on groups with smaller $\fractaldim$ and $\modularity$ values demonstrate better generalization to R3 and PS, respectively. This suggests that training on problems with similar structural values could enhance GNN generalisation to the target domain.

However, this pattern does not hold for certain properties and domains, such as $\modularity$ in AM, where AM has a larger $\modularity$ than SR, yet models trained on a smaller subset of SR generalize better to AM. One possible explanation for this discrepancy is that GNN performance is not determined by a single structural property but by a combination of structural features. Another possibility is that numerical structural values alone may not be sufficient to fully capture the problem structures learned by the GNN. Additionally, for domains like AM, while altering the $\scalefreev$ value in the training set does affect testing outcomes, even the improved models struggle to achieve strong performance, which needs further exploration to understand this phenomenon.

\begin{table}[!htbp]
        \centering
        \small
        \caption{Accuracy of models trained on SR(10-40) with different structural splits. }
        \vspace{2pt}
        \resizebox{0.4\textwidth}{!} {%
        \begin{tabular}
            {>{\centering\arraybackslash}p{0.46cm} >
            {\centering\arraybackslash}p{1cm} >
            {\centering\arraybackslash}p{0.65cm} >
            {\centering\arraybackslash}p{0.65cm} >{\centering\arraybackslash}p{0.65cm} >{\centering\arraybackslash}p{0.65cm} >{\centering\arraybackslash}p{0.65cm} >
            {\centering\arraybackslash}p{0.65cm} 
        }
        \toprule
           \multirow{3}{*}{Metric} &
           \multirow{3}{*}{Split} & \multicolumn{6}{c}{Testing Domains}        \\ \cmidrule(lr) {3-8}
        &&\multicolumn{2}{c}{NeuroSAT} & \multicolumn{2}{c}{GIN}  & \multicolumn{2}{c}{GCN}\\ \midrule
        \multirow{2}{*}{$\fractaldim$} & small & \multicolumn{2}{c}{$66.4\pm0.7$} & \multicolumn{2}{c}{$68.5 \pm 1.5$} & \multicolumn{2}{c}{$58.6\pm1.3$}  \\ & big & \multicolumn{2}{c}{$68.1\pm1.2$} & \multicolumn{2}{c}{$64.8\pm2.1$} & \multicolumn{2}{c}{$57.0\pm0.4$} \\ \midrule
        \multirow{2}{*}{$\scalefreev$} & small & \multicolumn{2}{c}{$67.5\pm1.3$} & \multicolumn{2}{c}{$63.4\pm1.1$} & \multicolumn{2}{c}{$61.5 \pm 0.6$}  \\  & big & \multicolumn{2}{c}{$67.8 \pm 2.1$} & \multicolumn{2}{c}{$67.1 \pm 2.3$} & \multicolumn{2}{c}{$61.5\pm0.5$} \\ \midrule
        \multirow{2}{*}{$\scalefreec$} & small& \multicolumn{2}{c}{$68.8\pm1.2$} & \multicolumn{2}{c}{$68.9 \pm 1.2$} & \multicolumn{2}{c}{$62.1 \pm 0.6$} \\ & big & \multicolumn{2}{c}{$68.0\pm3.2$} & \multicolumn{2}{c}{$65.1\pm1.8$} & \multicolumn{2}{c}{$62.0\pm1.0$} \\  \midrule
        \multirow{2}{*}{$\treewidth$} & small & \multicolumn{2}{c}{$66.8\pm0.9$} & \multicolumn{2}{c}{$66.7\pm3.5$}  &\multicolumn{2}{c}{$58.2\pm0.1$}  \\ & big & \multicolumn{2}{c}{$64.5\pm1.4$} & \multicolumn{2}{c}{$60.9\pm4.6$} & \multicolumn{2}{c}{$59.1\pm1.3$}  \\ \midrule
        \multirow{2}{*}{$\modularity$} & small & \multicolumn{2}{c}{$66.8 \pm2.5$} & \multicolumn{2}{c}{$64.5\pm2.5$} &\multicolumn{2}{c}{$60.1\pm0.1$}\\ & big & \multicolumn{2}{c}{$66.7\pm1.4$} & \multicolumn{2}{c}{$64.9\pm1.5$} &\multicolumn{2}{c}{$59.8\pm0.5$} \\  \midrule
        \multirow{2}{*}{$\Entropy$} & small & \multicolumn{2}{c}{$68.5\pm2.0$} & \multicolumn{2}{c}{$67.4\pm0.9$} & \multicolumn{2}{c}{$63.6\pm2.9$}\\
        & big & \multicolumn{2}{c}{$66.2\pm 4.4$} & \multicolumn{2}{c}{$63.3 \pm 3.3$} & \multicolumn{2}{c}{$56.9 \pm 0.9$}  \\ 
    \bottomrule
    \end{tabular}
    }
        \vspace{-0.2cm}
        \label{tab:general}
\end{table}

\begin{table}[t!]
        \centering
        \normalsize
        \caption{Average accuracy of GIN trained on SR(10-40) with different structural splits.}
        \vspace{2pt}
        \resizebox{0.5\textwidth}{!}{\begin{tabular}{>{\centering\arraybackslash}p{0.46cm} >
        {\centering\arraybackslash}p{1cm} >
        {\centering\arraybackslash}p{0.65cm} >{\centering\arraybackslash}p{0.65cm} >{\centering\arraybackslash}p{0.65cm} >{\centering\arraybackslash}p{0.65cm} >
        {\centering\arraybackslash}p{0.65cm} >
        {\centering\arraybackslash}p{0.65cm} >
        {\centering\arraybackslash}p{0.65cm} >
        {\centering\arraybackslash}p{0.65cm} >
        {\centering\arraybackslash}p{0.65cm} >
        {\centering\arraybackslash}p{0.65cm}}
        \toprule
           \multirow{3}{*}{Metric} &
           \multirow{3}{*}{Split} & \multicolumn{10}{c}{Testing Domains}        \\ \cmidrule(lr) {3-12}
        &&R3 & KCL & KD  & KV & KCO & AM & CA & PS & G2 & IN \\ \midrule
        \multirow{2}{*}{$\fractaldim$} & small & 92.0 & 53.3 & 64.6& 65.0 & 49.0 & 50.4 & 75.7 & 91.5 & 47.2 & 86.1\\ & big & 88.3 & 57.2 & 62.5 & 59.2 & 52.2 & 46.8 & 64.6 & 88.1 & 61.1 & 63.9\\ \midrule
        \multirow{2}{*}{$\scalefreev$} & small & 93.5& 51.6& 55.3& 55.9& 47.6&47.4&65.5 &93.8&50.0&80.6\\  & big & 92.3& 53.4&59.0&65.5&50.0&52.3&70.5&93.7&50.0&63.9\\ \midrule
        \multirow{2}{*}{$\scalefreec$} & small& 93.3 & 54.3& 60.1& 65.5& 59.6& 50.3& 68.0& 95.4& 58.3& 79.2\\ & big &93.4&51.7&54.8&59.0&56.7&49.0&65.0&92.1&52.8&57.0 \\  \midrule
        \multirow{2}{*}{$\modularity$} & small &90.8&48.8&59.9&60.8&49.4 & 53.6&63.6&91.6&50.0&86.1\\ & big & 89.9&53.2&63.2&62.5&50.0& 52.7& 65.6&86.9&38.7& 68.1\\  \midrule

    \bottomrule
    \end{tabular}}
        \vspace{-0.2cm}
        \label{tab:general2}
\end{table}

\begin{table} [!htbp]
    \caption{ Augmented experiments. Top: NeuroSAT trained on augmented datasets. Testing dataset is split to augmented and raw. Bottom: Average graph based properties before and after augmenting.}
    \begin{subtable}[!htbp]{0.50\textwidth}
    \small
    \centering 
    \normalsize
    \resizebox{1\columnwidth}{!}{%
    \begin{tabular}{>{\centering\arraybackslash}p{0.46cm} >
        {\centering\arraybackslash}p{1.5cm} >
        {\centering\arraybackslash}p{0.75cm} >
        {\centering\arraybackslash}p{0.65cm} >{\centering\arraybackslash}p{0.65cm} >{\centering\arraybackslash}p{0.65cm} >
        {\centering\arraybackslash}p{0.66cm} >
        {\centering\arraybackslash}p{0.65cm} >
        {\centering\arraybackslash}p{0.65cm} >
        {\centering\arraybackslash}p{0.65cm} >
        {\centering\arraybackslash}p{0.65cm}}
        \toprule
        Domain & Split & R3 & KCL & KV & KD & KCO & CA & PS & G2 & IN \\ \midrule
         \multirow{2}{*}{SR} & augmented  & \textbf{99.9} & 50.0 & \textbf{58.9}& \textbf{64.0}& \textbf{94.4}&\textbf{70.4} & \textbf{99.6 }& 4.2 &\textbf{66.7}\\ & raw & 50.0 & 50.0 & 56.0& 54.0& 57.4& 50.0 & 51.1 & 4.2& 41.7\\  \midrule
    \end{tabular} }
    \label{tab:augment}
    
    \end{subtable} %

    \vspace{2.2pt}

    \begin{subtable}[!htbp]{0.5\textwidth}
    \centering
    \normalsize
    \small
    \resizebox{0.9\columnwidth}{!}{%
    \begin{tabular}{>{\centering\arraybackslash}p{0.46cm} >
        {\centering\arraybackslash}p{1.5cm} >
        {\centering\arraybackslash}p{0.75cm} >
        {\centering\arraybackslash}p{0.65cm} >{\centering\arraybackslash}p{0.65cm} >{\centering\arraybackslash}p{0.65cm} >{\centering\arraybackslash}p{0.65cm} >
        {\centering\arraybackslash}p{0.65cm} >
        {\centering\arraybackslash}p{0.65cm}}
        \toprule
        Domain & Split  & $\fractaldim$ & $\scalefreec$ & $\scalefreev$ & $\modularity$ &$\treewidth$ & $\centralityb$  & $\Entropy$  \\ \midrule
         \multirow{2}{*}{SR}& augmented & 2.5 & 4.2 & 9.8 & 0.4 & 39.2 &0.01&5.0 \\ & raw & 3.4 & 4.3 & 10.0 & 0.4 & 39.1 & 0.01 & 5.0 \\  \midrule
    \end{tabular}
    \label{tab:augment1}
    }
    
    \end{subtable} %

            \vspace{-0.6cm}
    \label{tab:totalgraph}    
\end{table}

\paragraph{Augmented problem generalisation.}
To get insights to Q4, we augment SR(10-40) with learned clauses from Cadical solver. To do this, we append all the learned clauses to the ``raw" problems in chronological order in which they were generated by Cadical, and label them as ``augmented" set in the experiment. Then, we test the results on different domains as shown in \cref{tab:totalgraph}. Both training and testing data here are not splitted with structural properties. Although models trained on augmented datasets reach high accuracy in other augmented domains, with only a few learned clauses added (on average the ratio between learned clause and original clause is roughly 15.9\% in training data), they could not solve any raw data well in any domains. While most structural parameters remain the same before and after augmentation, the relatively large change in $\fractaldim$ could be one reason for this result.

\subsection{Analysis}
The results in both \cref{tab:experiment} and \cref{sec:appendix_result} suggest that the generalisability of GNN is influenced by the structure of the data to varying extent. For CNF-based properties, this influence is most obvious. GNNs tend to find certain types of problems, such as those with a $c/v$ ratio at the phase transition or lower backbone sizes, more challenging compared to others under the same evaluation metric. For graph-based properties, the influence is less pronounced. 
For some results in \cref{tab:general2}, the accuracy differences between the small and large metric splits are not significant across all models, which indicates one of two possibilities: (1) all the models fail to capture the impact of the metrics, or (2) the metrics within the training set have minimal influence on model accuracy. In this work, we adopt the second assumption since we use three different models, and their performance aligns with this interpretation. However, we acknowledge that the first assumption could also hold, which should be analyzed in future work. Furthermore, if one model shows significant differences while another does not, we infer that the less significant model fails to effectively capture the impact of the property in the training set. 
Overall, while some structural variations are not effectively captured by GNNs, as evidenced across all three models, there is a general trend of GNNs performing better on problems with similar graph structures. However, this tendency does not hold for certain problem types, highlighting the need for further investigation into combined structural properties. 

\section{Conclusion}
\label{conclusion}
This paper presents StructureSAT, a structure-based SAT dataset with various domains and property measures. We aim to help analyze GNN based generalisation in SAT solving by adapting a variety of structural properties. Through our extensive cross-domain experiments using StructureSAT, we produced various insights in GNN generalisation with empirical analysis. As suggested from our experiments, certain properties are more influential on generalisability than other properties. Furthermore, the diverse GNN generalisation abilities on different domains could be the result of a combination of properties. We hope StructureSAT brings interest into the role of structural properties for future research into GNN based SAT solving. 

\bibliography{cite}
\clearpage

\begin{appendix}
\section{Limitations and Future Work}
\label{sec:appendix_limitation}
Despite being the first dataset to more deeply analyse SAT structure in terms of GNN generalisation, there are still a few limitations and many possibilities to consider for future work. Firstly, StructureSAT is separated into subsets, where each subset has one domain and focuses on one type of feature at a time. While useful for identifying individual factors of influence at a property level, a multivariate analysis of features and domains could be considered in future work. Secondly, due to certain problems being fundamentally intractable, we do not consider some other CNF based features. The structure calculations in the distribution plots are limited to a 60 seconds timeout as most industrial and G2SAT generative processes are computationally hard to compute. Exploring effective algorithms of computing the structural properties on SAT, especially on large problems, could be a potential future direction. Thirdly, although we include SATCOMP2007 in our dataset and calculate structural values of individual problems, the models we trained could not be efficiently tested on large industrial and G2SAT problems. Thus we only test on 24 G2SAT problems and 12 small industrial problems used to generate the G2SAT problems inside the experiments. However, either better models or more efficient encoding strategies could be included in future work for GNNs to solve large industrial problems. Furthermore, to simplify the comparison process between training and testing structures, we ignored the possible structure changes and losses during the process of encoding and embedding the SAT problems, which should be considered and analysed in future work. Lastly, as we generated 200k pairs of data for each domain and divided the base dataset, there is overlap between different property split groups. Potential future work could be generating datasets with more controllable feature values as the CA generator does \cite{giraldez-cruModularityBasedRandomSAT2015}. 

\section{Full analysis on dataset }
\label{fulldata}
\subsection{Distribution plot}
To get a comprehensive understanding of base dataset domains without splitting, we extend previous experiments and plot 6 other graph property distributions including $\scalefreec$, $\Entropy$, $BE$, $\Proximaty$, $\treewidth$ and $c/v$ ratio. Since LCG is a bipartite graph, some calculations have been adjusted to accommodate its bipartite characteristics. For example, the clustering coefficient in $\Proximaty$ has been modified to use the bipartite clustering coefficient calculation.

\begin{figure*} [!htb]
    \centering
    \includegraphics[width=0.9\textwidth]{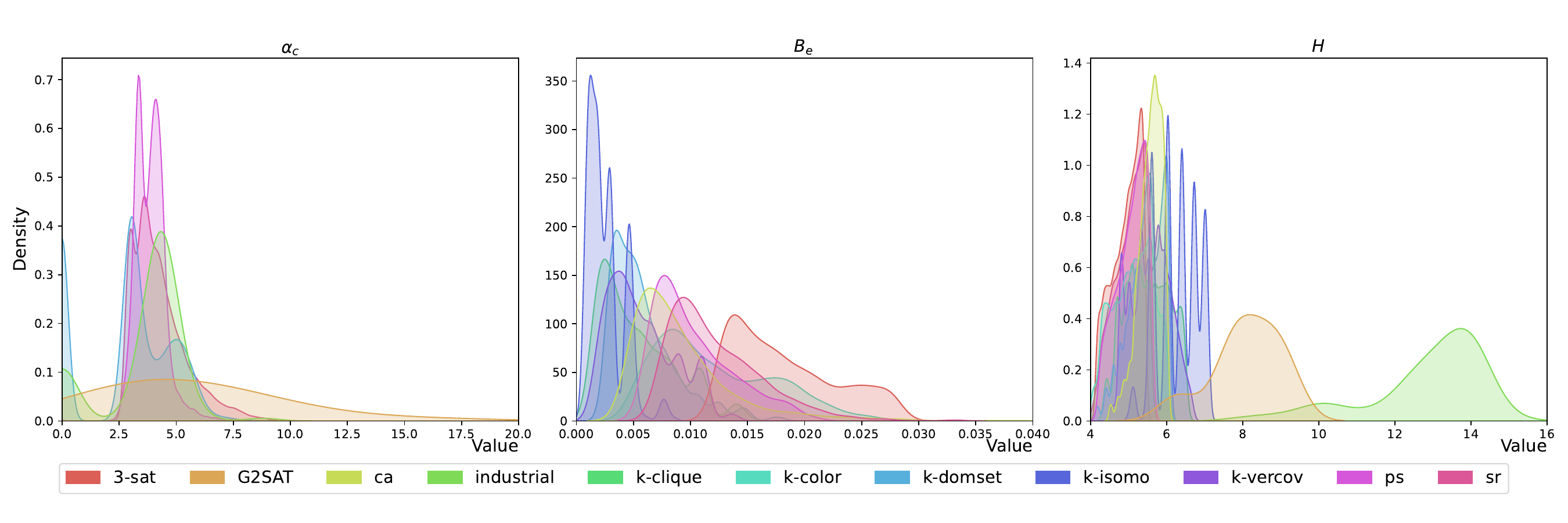}
    \vspace{-5pt}
    \caption{Distribution of various structural graph properties from different problem domains. Left: scale-free measure by clause $\scalefreec$. Middle: centrality measure$BE$. Right: entropy measure $\Entropy$.}
    \label{fig:metric1}
    \vspace{-10pt}
\end{figure*}

\begin{figure*} [!htb]
    \centering
    \includegraphics[width=0.9\textwidth]{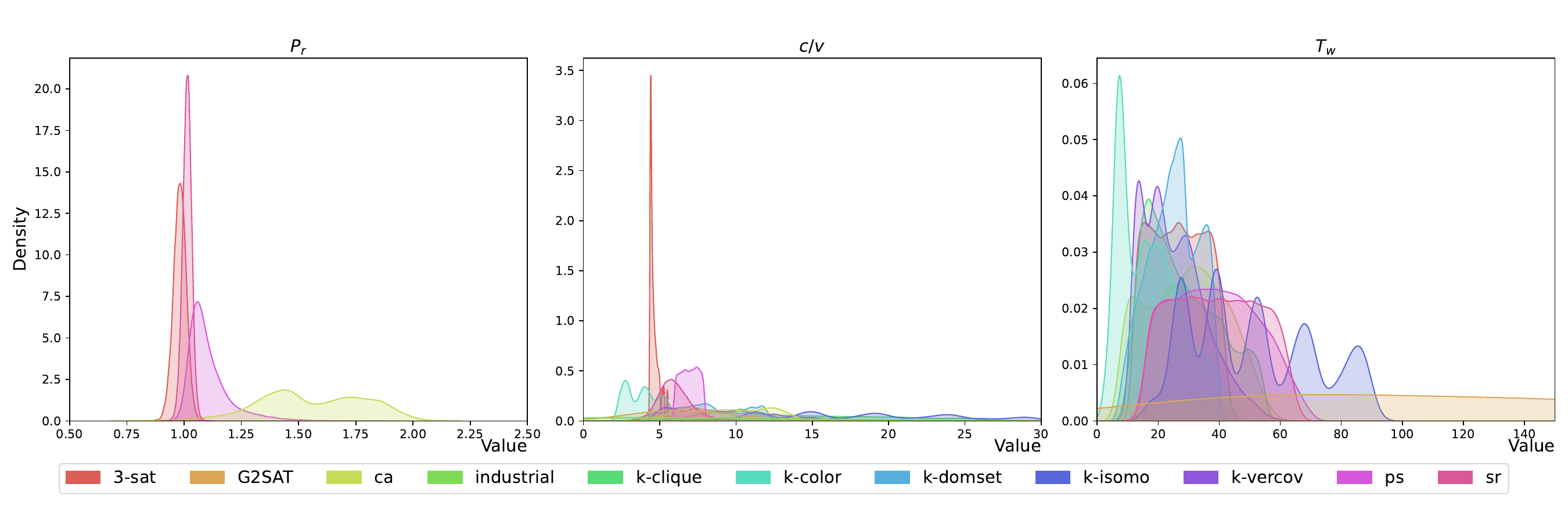}
    \vspace{-5pt}
    \caption{Distribution of various stuctural graph properties from different problem domains. Left: Proximity:$\Proximaty$. Middle: ratio $c/v$. Right: Treediwth $\treewidth$. }
    \label{fig:metric2}
    \vspace{-10pt}
\end{figure*}

The figures show a wide variety of distributions over graph properties among different domains. Some distinguishable differences can be seen within $\modularity$, where random, crafted, and industrial problems have low, medium, and high values, respectively. Additionally, industrial and crafted domains have a wider range of $\fractaldim$ values than random domains. In \cref{fig:metric1}, the value range of $B_e$ ranks from lowest to highest in crafted, industrial, and random domains, while industrial problems have highest $\Entropy$ values. For available $P_r$ values in \cref{fig:metric2}, the values from random domains 3-SAT and SR are close to 1, while the pseudo-industrial problems CA and PS have higher $P_r$. While SR has been experimented to be the domain with the best OOD generalisation ability among all the mentioned domains, it does not have the widest coverage of property values. E.g., SR domain is narrower in $H$ compared with other domains. All the distinct characteristics within the dataset highlight the potential abilities of affecting generalisation differently within GNNs.

\subsection{Structural plots for each domain}
\cref{tab:average11} contains average structure values of the used testing set from $D_{t3}$. We also show the average structure values used for splitting in the training and validation sets in \cref{tab:average111}. Note that this is the average value of all data in the raw datasets, which includes both Satisfiable (SAT) and Unsatisfiable (UNSAT) problems. However as there are differences in SAT and UNSAT problems of each domain due to the inherent randomness of the generation process, the average values between SAT and UNSAT are slightly different. Although the average values do not represent the overall structure and distribution for each domain, it is an estimation of the calculated structure for the problems and is used as the main metric in our experiments.

\cref{fig:srcore,fig:srcore1,fig:srcore2,fig:srcore3,fig:srcore4,fig:srcore5,fig:srcore6} present the correlation relationships between each of the calculated graph based properties, including number of variables and number of clauses, from the raw 200k datasets for each domain (datasets without splits). Each domain contains unique sturcture and different relationship patterns with the other structures, and some of those patterns are consistent across all domains. E.g., higher number of clauses always results in a lower entropy value. Some of the patterns are consistent with GNN generalisation results while others show opposite. For instance, when looking at the SR domain, higher $\treewidth$ values results in higher $\Entropy$ values, which is consistent with the pattern that both higher value of $\treewidth$ and $\Entropy$ results in better overall OOD results. However, while higher $\treewidth$ would also results in generally lower $\fractaldim$, higher valued $D_f$ group reaches better results on NeuroSAT.

\begin{table}[t]
    \centering
    \normalsize
    \vspace{-0.5cm}
    \caption{Mean structural properties of StructureSAT testing domains.
    N/A indicates unavailable.}
    \resizebox{0.5\textwidth}{!}{
    \begin{footnotesize}
    \begin{tabular}{lcccccccc}
        \toprule
        & $\fractaldim$ & $\scalefreev$ & $\scalefreec$ & $\treewidth$ &  $\centralityb$&$\modularity$ & $\Entropy$ \\ 
        \midrule
        SR (10-40) & 3.39 & 4.25 & 10.03 & 39.12 & 0.012 
        & 0.38 & 5.01 \\ 
        \midrule
        r-3SAT & 2.57 & 7.54 & N/A & 26.53 & 0.018 
        & 0.45 & 4.92 \\ 
        \midrule
        $k$-clique & 3.67 & 13.44 & N/A & 28.19 & 0.005 
        & 0.47 & 5.51 \\ 
        \midrule
        $k$-dominating-set & 2.88 & 5.44 & 2.95 & 25.03 & 0.006 
        & 0.49 & 5.46 \\ 
        \midrule
        $k$-vertex-cover & 2.89 & 6.29 & N/A & 25.24& 0.006 
        & 0.50 & 5.58 \\ 
        \midrule
        $k$-coloring & 2.00 & 2.92 & N/A & 17.73 & 0.02 
        & 0.64 &  5.13 \\ 
        \midrule
        automorph & 3.51 & 17.47 & N/A &  51.25 & 0.0026
        & 0.49 & 6.28 \\ 
        \midrule
        CA & 2.92 & 12.13 & N/A & 29.39 & 0.009  
        & 0.73 & 5.56 \\ 
        \midrule
        PS & 3.27 & 5.49 & 3.87 & 38.66 & 0.01 &
         0.36 &  5.06 \\ 
        \midrule
        G2SAT & 3.06 & 5.30 & 5.56 & 116.87 & N/A &
        0.87 & 8.09 \\ 
        \midrule
        industrial & 2.68 & 10.09 & 3.50 & N/A
        & N/A & 0.87 & 12.49 \\ 
        \bottomrule 
    \end{tabular}
    \end{footnotesize}}
    \vspace{-0.5cm}
    \label{tab:average111}
\end{table}

\begin{table}[t]
    \centering
    \normalsize
    \vspace{-0.5cm}
    \caption{Mean structural properties of StructureSAT raw domains.
    N/A indicates unavailable.}
    \resizebox{0.5\textwidth}{!}{
    \begin{footnotesize}
    \begin{tabular}{lcccccccc}
        \toprule
        & $\fractaldim$ & $\scalefreev$ & $\scalefreec$ & $\treewidth$ &  $\centralityb$&$\modularity$ & $\Entropy$ \\ 
        \midrule
        SR (10-40) & 3.00 & 4.25 & 10.03 & 39.23 & 0.01 
        & 0.38 & 5.01 \\ 
        \midrule
        r-3SAT & 2.57 & 7.52 & N/A & 26.53 & 0.02 
        & 0.45 & 4.92 \\ 
        \midrule
        $k$-clique & 3.26 & 14.23 & N/A & 29.93 & 0.005 
        & 0.47 & 5.61 \\ 
        \midrule
        $k$-dominating-set & 2.89 & 5.44 & 3.01 & 25.33 & 0.01 
        & 0.49 & 5.48 \\ 
        \midrule
        $k$-vertex-cover & 2.91 & 6.29 & N/A & 25.47& 0.01 
        & 0.50 & 5.60 \\ 
        \midrule
        $k$-coloring & 2.38 & 2.92 & N/A & 17.73 & 0.02 
        & 0.64 &  5.56 \\ 
        \midrule
        CA & 2.92 & 12.16 & N/A & 29.36 & 0.01  
        & 0.73 & 5.56 \\ 
        \midrule
        PS & 3.27 & 5.49 & 3.87 & 38.71 & 0.01 &
         0.36 &  5.06 \\ 
        \bottomrule 
    \end{tabular}
    \end{footnotesize}}
    \vspace{-0.5cm}
    \label{tab:average11}
\end{table}

\clearpage

\begin{figure*} [!htb]
    \centering
    \includegraphics[width=0.95\textwidth]{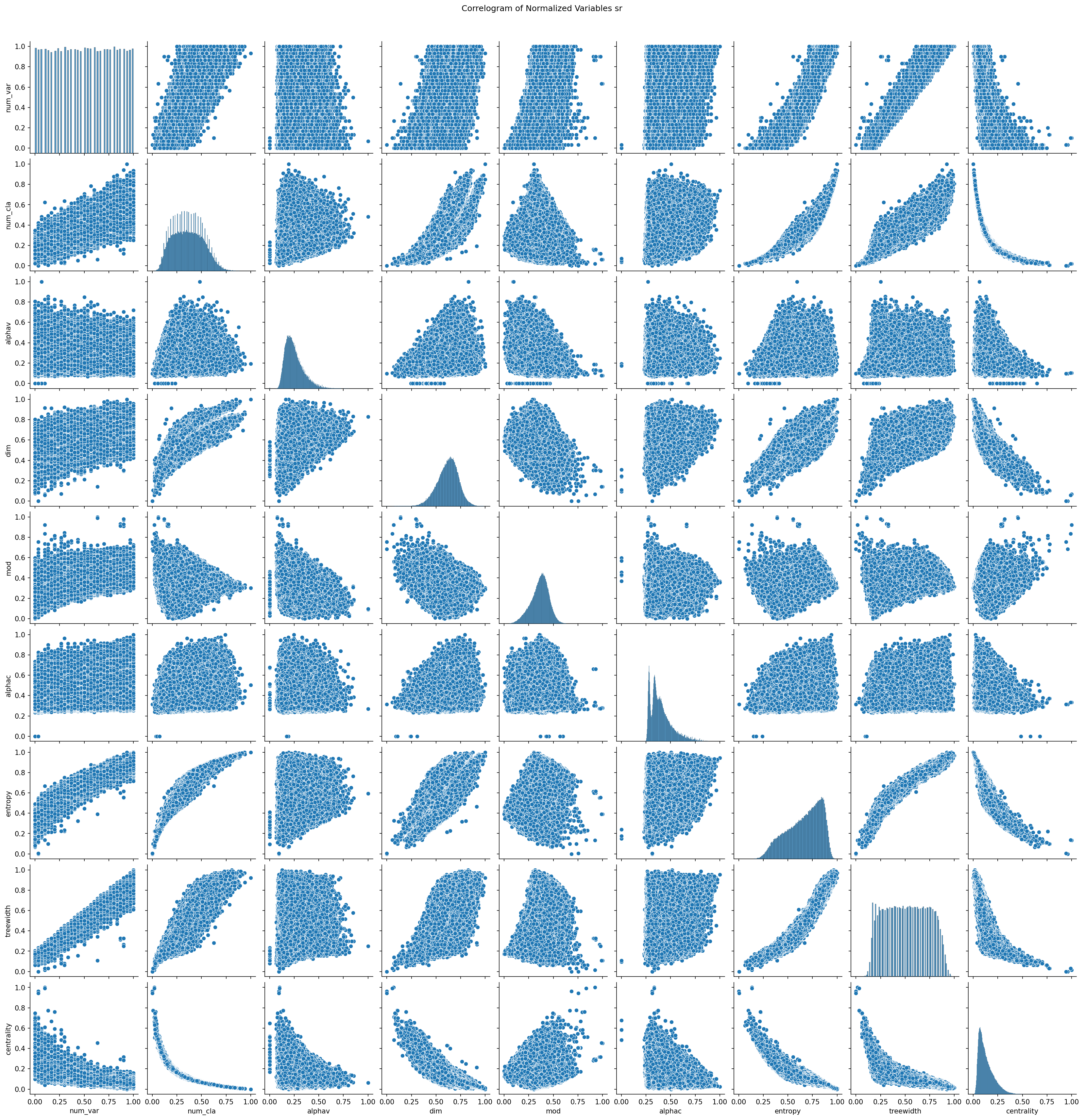}
    \vspace{-5pt}
    \caption{SR correlogram. }
    \label{fig:srcoresr}
    \vspace{-10pt}
\end{figure*}

\begin{figure*} [!htb]
    \centering
    \includegraphics[width=0.95\textwidth]{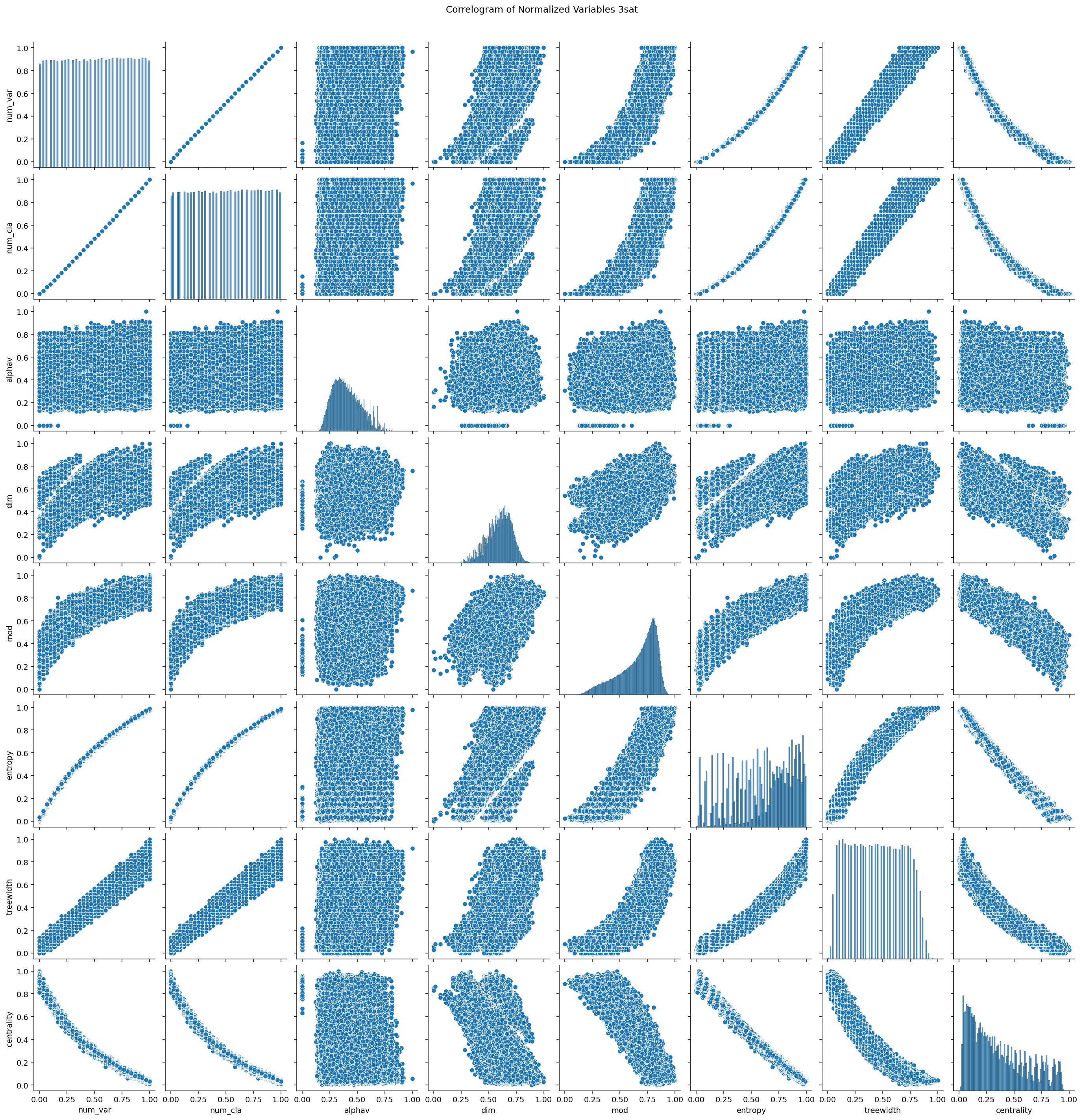}
    \vspace{-5pt}
    \caption{Random 3-SAT correlogram. }
    \label{fig:srcore1}
    \vspace{-10pt}
\end{figure*}

\begin{figure*} [!htb]
    \centering
    \includegraphics[width=0.95\textwidth]{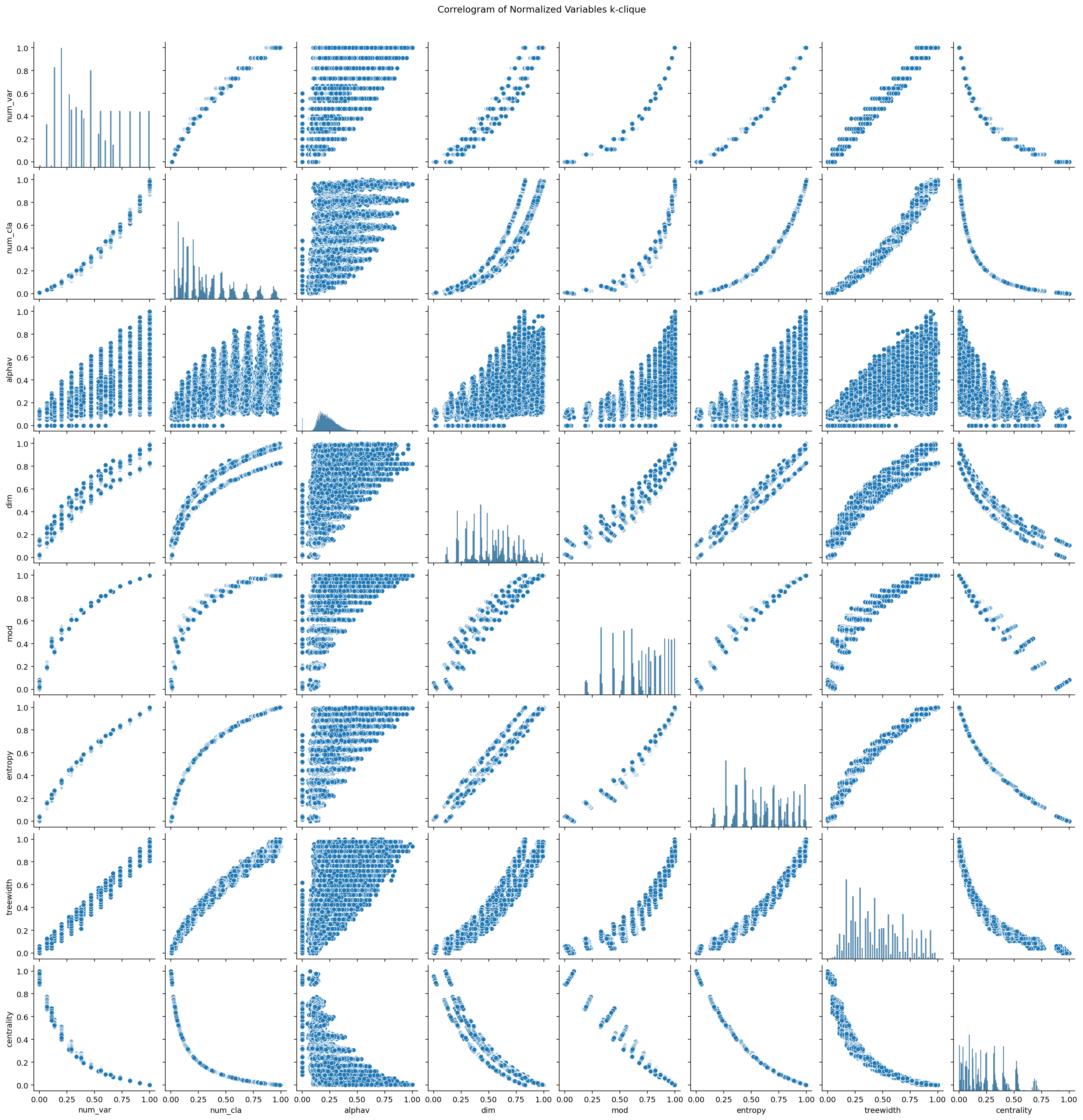}
    \vspace{-5pt}
    \caption{k-clique correlogram. }
    \label{fig:srcore2}
    \vspace{-10pt}
\end{figure*}

\begin{figure*} [!htb]
    \centering
    \includegraphics[width=0.95\textwidth]{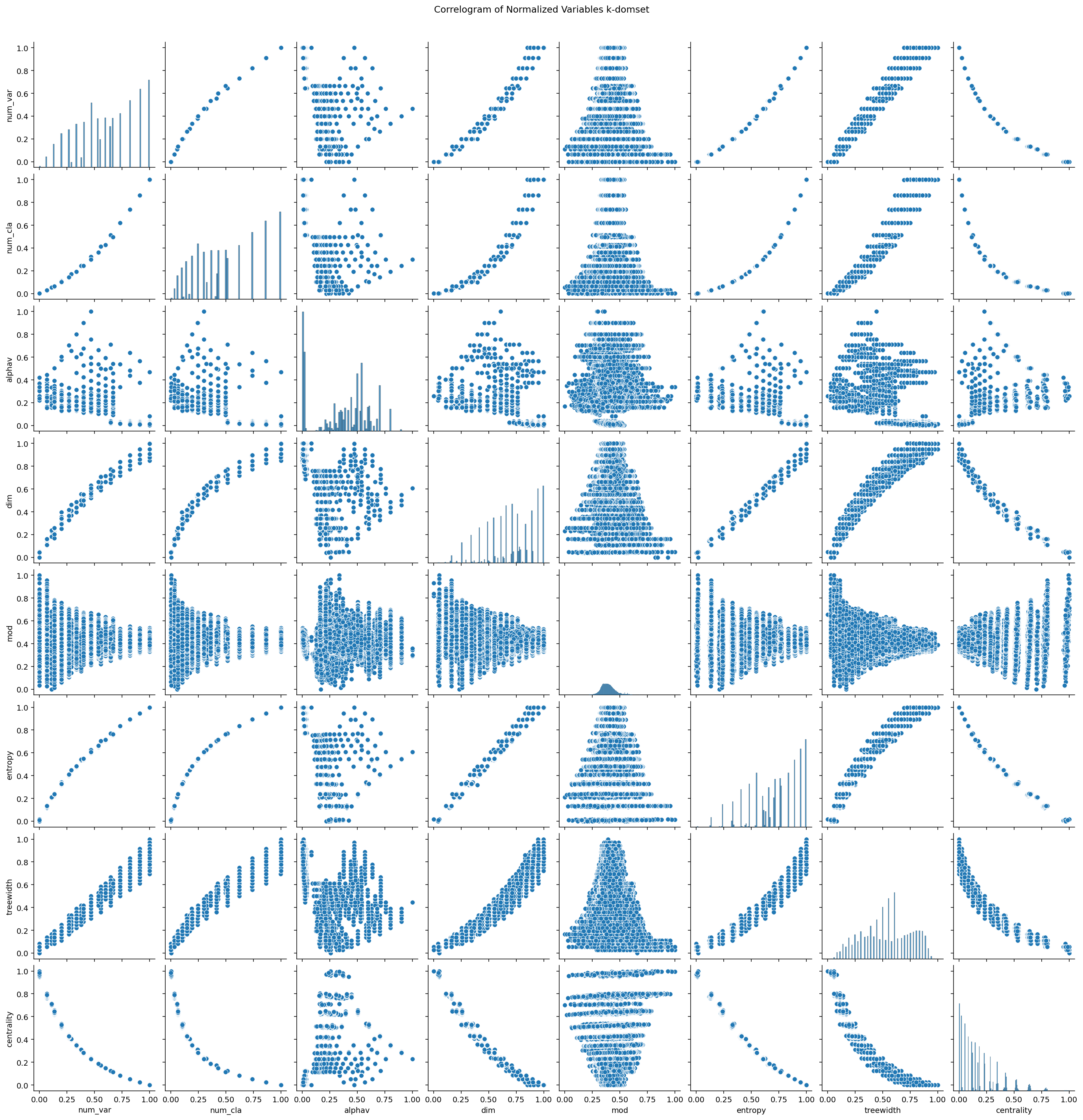}
    \vspace{-5pt}
    \caption{k-domset correlogram. }
    \label{fig:srcore}
    \vspace{-10pt}
\end{figure*}

\begin{figure*} [!htb]
    \centering
    \includegraphics[width=0.95\textwidth]{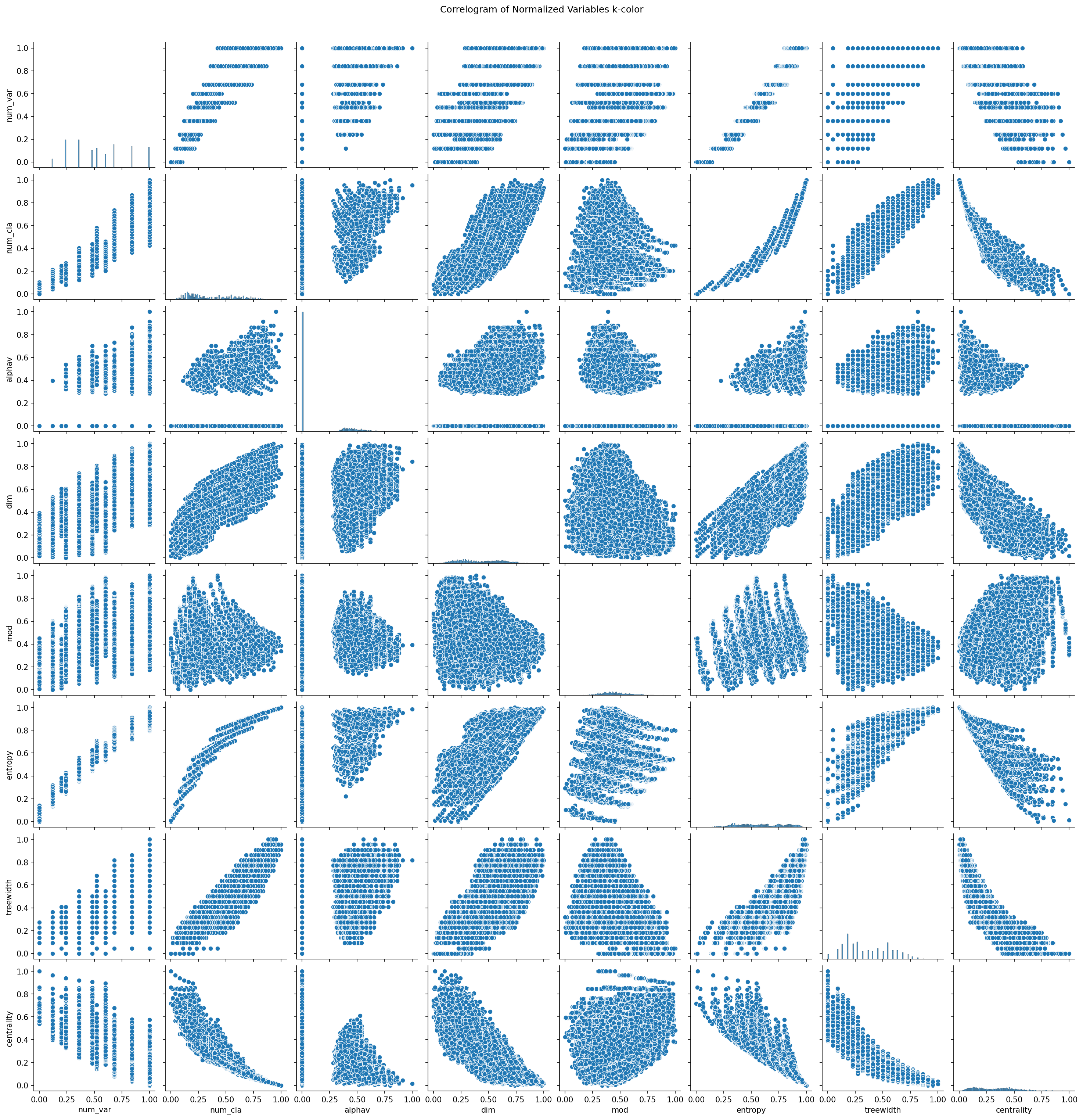}
    \vspace{-5pt}
    \caption{k-color correlogram. }
    \label{fig:srcore3}
    \vspace{-10pt}
\end{figure*}

\begin{figure*} [!htb]
    \centering
    \includegraphics[width=0.95\textwidth]{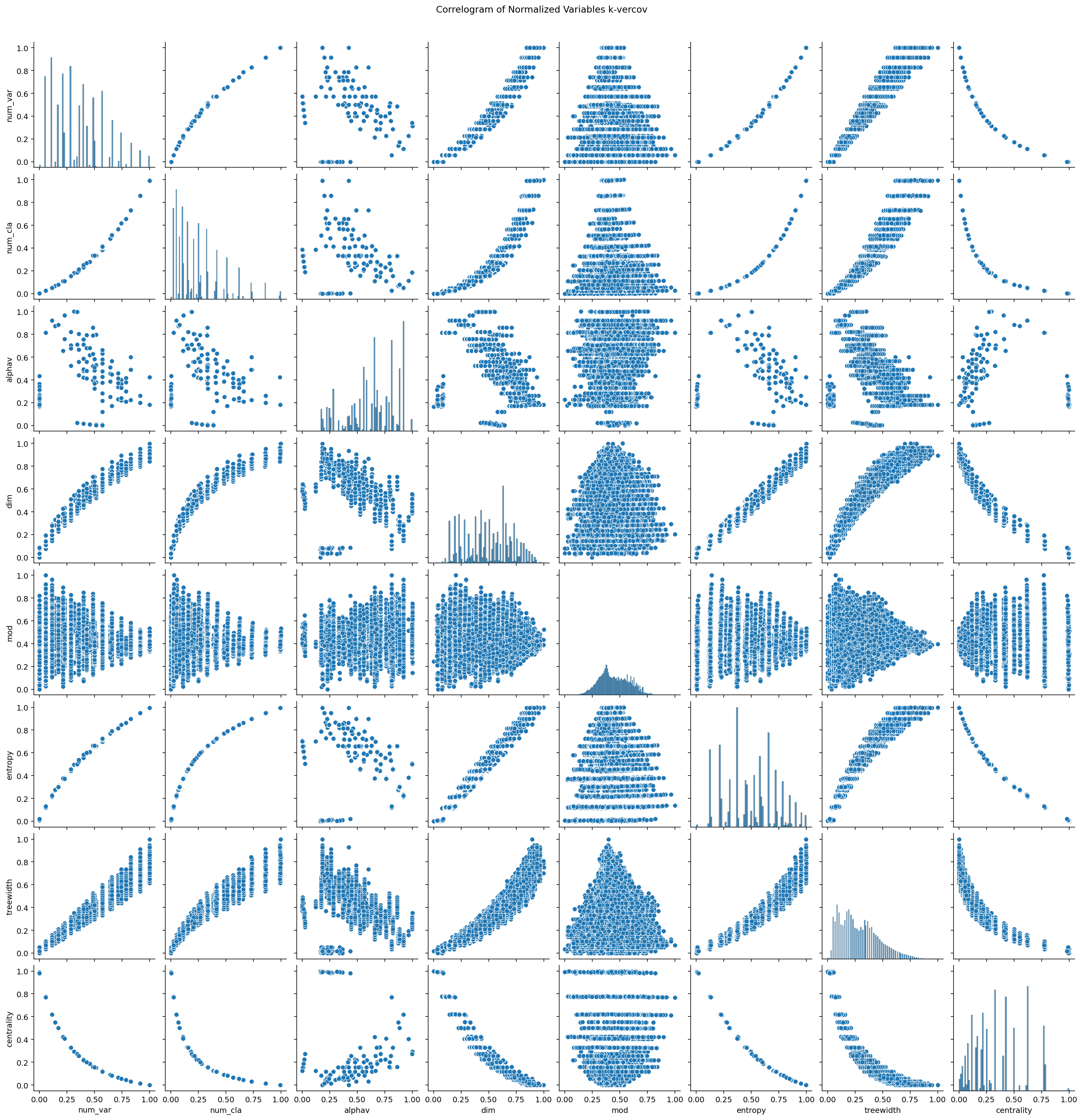}
    \vspace{-5pt}
    \caption{k-vercov correlogram. }
    \label{fig:srcore4}
    \vspace{-10pt}
\end{figure*}

\begin{figure*} [!htb]
    \centering
    \includegraphics[width=0.95\textwidth]{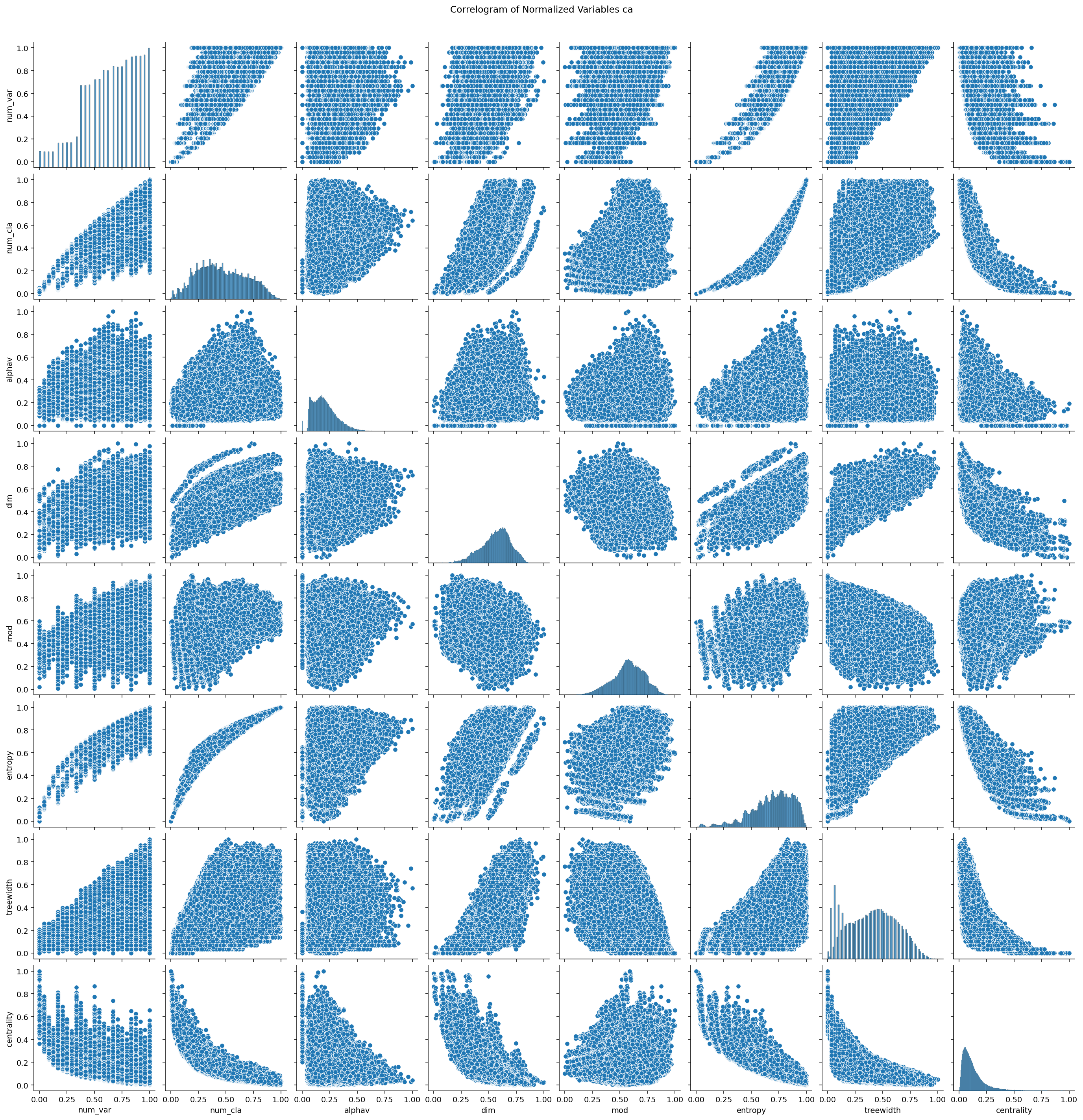}
    \vspace{-5pt}
    \caption{ca correlogram. }
    \label{fig:srcore5}
    \vspace{-10pt}
\end{figure*}

\begin{figure*} [!htb]
    \centering
    \includegraphics[width=0.95\textwidth]{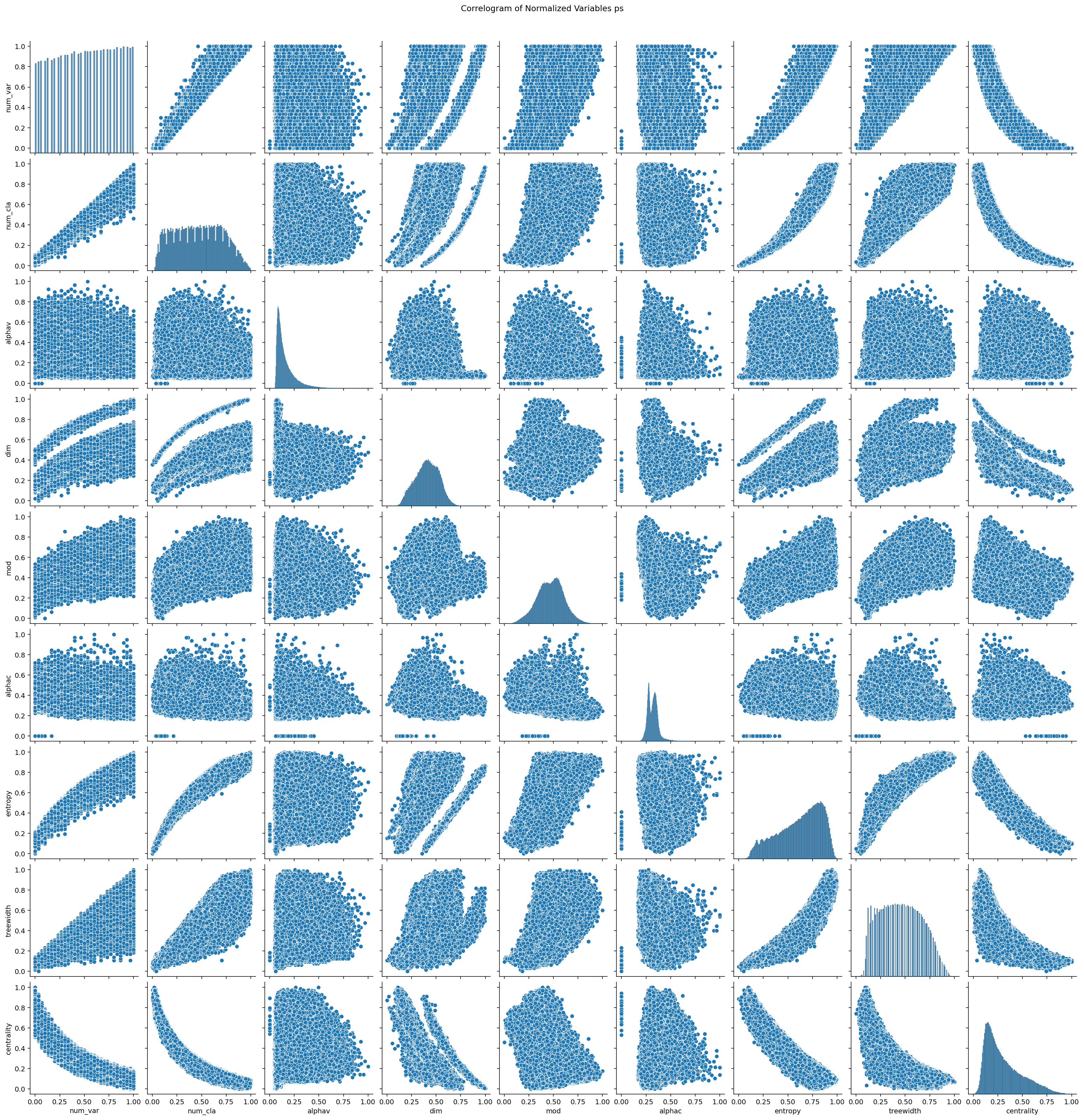}
    \vspace{-5pt}
    \caption{ps correlogram. }
    \label{fig:srcore6}
    \vspace{-10pt}
\end{figure*}

\clearpage
\section{Additional experimental details}
\label{sec:more_exp_details}

\subsection{GNN baseline}
\label{sec:appendix_GNN}
Models used in this work include NeuroSAT \cite{selsam2018learning}, GCN \cite{kipf2016semi} and GIN \cite{xu2018powerful}. 
\subsection{Code Base, experiment set up, and license}
\label{sec:appendix_helpsetup}
Our Dataset and Code base are available in\footnote{Our dataset and codebase are available \href{https://drive.google.com/drive/folders/1ZrhRlRqQUTrYExVzVbXaocjvNIi2Os25?usp=sharing.}{here}.\\}. The link also include detailed instruction on downloading and running experiment with the dataset.

Experiments with GNN are done using existing work from \cite{liG4SATBenchBenchmarkingAdvancing2023a}. Experimental parameters include $1e-04$ learning rate, $1e-08$ and $1e-07$ weight decay and $32$ number of message passing iterations. Batch sizes are selected from {128, 32, 16}. All the experiments are run on a machine with a NVIDIA A4500. 

StructureSAT is openly licensed via \href{https://creativecommons.org/licenses/by/4.0/}{CC BY 4.0}.

\section{Additional experimental results}
This section details some additional experimental results and analysis.
\label{sec:appendix_result}
\subsection{In-domain generalisation}

\subsubsection{In-domain generalisation to larger problem}
In \cref{tab:large3sat}, we train NeuroSAT models using R3, then tested on larger problems with more number of variables(40-100, 100-200). All problems are generated with the ratio $c/v$ at the phase transition.  In general, in both medium sized testing sets and large sized testing sets, the bigger splitted groups perform better. This matches with the pattern shown in Figure 4, where larger number of variables results in a larger value in almost all structures focused in \cref{tab:large3sat}. Thus, from the results we get in the in-domain generalisation experiments so far, we could conclude that for random problem domains, the change in structure values in training set could results in different testing performance to larger problems, where bigger training domain values are almost always better.

\begin{table}[t]
        \centering
        \normalsize
        \caption{{NeuroSAT testing results. 
        For each metric, we train NeuroSAT on the small and big split of the corresponding R3 training sets. 
        We then test the trained model on the testing data of 10-40, 40-100, and 100-200 variables for each metric. 
        Better training split performance in bold.}}
        \vspace{2.7pt}
        \resizebox{0.48\textwidth}{!}{
        \begin{footnotesize}
        \begin{tabular}{>{\centering\arraybackslash}p{1.2cm} >
        {\centering\arraybackslash}p{2.4cm} >
        {\centering\arraybackslash}p{0.75cm} >
        {\centering\arraybackslash}p{0.85cm} >{\centering\arraybackslash}p{0.85cm} >{\centering\arraybackslash}p{0.85cm} >{\centering\arraybackslash}p{0.85cm}}
        \toprule
           \multirow{2}{*}{Train Split} &
           \multirow{2}{*}{Test Set} &
           \multicolumn{5}{c}{Metric}       \\ \cmidrule(lr) {3-7}
        && $\fractaldim$ & $\scalefreev$ & $\treewidth$& $\modularity$ &  $\Entropy$\\ \midrule
          
         small & \multirow{2}{*}{40-100} &  78.0 & 81.0 & 78.9 & 66.3 & 72.0\\ big & & 82.0&83.0 & 81.4 & 81.4 & 81.7 \\ \midrule
         small &  \multirow{2}{*}{100-200} &70.7 & 76.0 
         & 72.6 & 54.4 & 60.2\\  \centering big & &75.2& 75.2& 74.1 & 72.7 & 74.9\\ \midrule
    \bottomrule
    \end{tabular}
    \end{footnotesize}}
        \vspace{-0.5cm}
        \label{tab:large3sat}
\end{table}

\subsubsection{In-domain generalisation to different backbone sizes}

To examine the impact of backbone sizes on the generalizability of Graph Neural Networks (GNNs), we trained models across various domains without structure splits and tested them on same domain problems with differing backbone sizes. \cref{tab:combined_results} presents the results for models trained and tested on SR and R3, respectively. The test sets comprise solely satisfiable problems from $D_{t3}$, categorized according to the average backbone size. For both domains on NeuroSAT, the performance is better on small backbone size problems than on bigger ones. To further check this phenomenon, tables \ref{tab:3sat11} and \ref{tab:3sat12} display the performance of NeuroSAT and GCN models trained on R3 without splits and tested on satisfiable problems with varying backbone sizes. These problems, sourced from SATLIB \cite{hoosSATLIBOnlineResource}, maintain a controlled number of variables (100) and clauses (403). The results indicate that, although the problem sizes are consistent, variations in backbone sizes significantly affect the accuracy of both models. The presence of more extensive backbones within a problem increases its complexity and the difficulty of finding a solution for GNNs.

Based on the results from $c/v$ ratio experiments, we conclude that while GNNs are given input with graph-structured data, meaning that they mainly operate on graph-based structures, their performance is also influenced by the CNF encoding of the domains. Therefore, using domain size as the sole metric for testing generalizability and assessing problem difficulty is limited, as there are small-sized yet challenging problems for GNNs to solve. These findings also suggest that changes in graph structure could result from alterations in CNF-based properties, which should be furthure tested.

\begin{table}[h!]
\centering
\normalsize
\vspace{-0.3cm}
\caption{Results with different backbone sizes on NeuroSAT. Top SR bottom R3.}
\begin{minipage}{0.45\textwidth}
    \centering
    \begin{footnotesize}
    \begin{tabular}{lcc}
    \toprule
    Metric & Accuracy \\
    \midrule
    big    & 89.5    \\
    small  & 98.2  \\
    \bottomrule
    \end{tabular}
    \end{footnotesize}
\end{minipage} \hfill
\begin{minipage}{0.45\textwidth}
    \centering
    \begin{footnotesize}
    \begin{tabular}{lcc}
    \toprule
    Metric & Accuracy \\
    \midrule
    big    & 74.7    \\
    small  & 95.0  \\
    \bottomrule
    \end{tabular}
    \end{footnotesize}
\end{minipage}
\vspace{-0.1cm}
\label{tab:combined_results}
\end{table}

\begin{table}
\centering
\normalsize
\vspace{-0.3cm}
\caption{R3 results with different backbone sizes on NeuroSAT.}
    \begin{footnotesize}
    \begin{tabular}{lccc}
    \toprule
    Metric & Accuracy \\
    \midrule
    10        & 97.1    \\
    30          & 84.4  \\
    50   &   71.8  \\
    70       &  67.9  \\
    90         & 42.6 \\
    \bottomrule
\end{tabular}
    \end{footnotesize}
\vspace{-0.1cm}
\label{tab:3sat11}
\end{table}

\begin{table}
\centering
\normalsize
\vspace{-0.3cm}
\caption{R3 results with different backbone sizes on GCN.}
    \begin{footnotesize}
    \begin{tabular}{lccc}
    \toprule
    Metric & Accuracy \\
    \midrule
    10        & 91.0    \\
    30          & 77.6  \\
    50   &   65.8  \\
    70       &  54.8  \\
    90         & 35.7 \\
    \bottomrule
\end{tabular}
    \end{footnotesize}
\vspace{-0.1cm}
\label{tab:3sat12}
\end{table}

\subsection{Out-of-domain generalisation}
In this section, we extend previous experiments on OOD generalisation test to other domains. Results are shown in \cref{tab:general3sat}. As discussed earlier, the general out-of-distribution testing performances vary depending on the training set, training structure focus and model used. The extend varies between structures as well, with $\modularity$ having a bigger difference influence on models trained using PS than using KCO overall. 
To some domains such as k-color, structural splits does little effect on the overall generalisation results across all three domains, while to domains like R3, $\scalefreev$ influences GIN more than GCN. Note that all experiments are run on a testing iteration of 32. Higher testing iterations could results in better generalisation, while the extent also varies depending on the model and domains. E.g., testing on a 200 number of iterations results in a 1\% increase in accuracy on KCO. Future work could be analysing the effect of higher iteration to the learning and embedding of the graph and its corresponding structure.

Furthermore, we trained NeuroSAT on SR(10-40) with different structural splits on satisfying assignment prediction task, which is considered a node classification task. Results in \cref{tab:general111} also indicate the importance of stucture while finding the correct assignment solution for a problem.

\begin{table}[!htbp]
        \centering
        \small
        \caption{Accuracy of models trained with different structural splits. }
        \vspace{2pt}
        \resizebox{0.4\textwidth}{!}
        {\begin{tabular}
        {>{\centering\arraybackslash}p{0.46cm} >
        {\centering\arraybackslash}p{1cm} >
        {\centering\arraybackslash}p{0.65cm} >
        {\centering\arraybackslash}p{0.65cm} >{\centering\arraybackslash}p{0.65cm} >{\centering\arraybackslash}p{0.65cm} >{\centering\arraybackslash}p{0.65cm} >
        {\centering\arraybackslash}p{0.65cm} 
}
        \toprule
           \multirow{3}{*}{\textbf{R3}} &
           \multirow{3}{*}{Split} & \multicolumn{6}{c}{Testing Domains}        \\ \cmidrule(lr) {3-8}
        &&\multicolumn{2}{c}{NeuroSAT} & \multicolumn{2}{c}{GIN}  & \multicolumn{2}{c}{GCN}\\ \midrule
        \multirow{2}{*}{$\scalefreev$} & small & \multicolumn{2}{c}{$67.5\pm1.3$} & \multicolumn{2}{c}{$63.4\pm1.1$} & \multicolumn{2}{c}{$61.5 \pm 0.6$}  \\  & big & \multicolumn{2}{c}{$67.8 \pm 2.1$} & \multicolumn{2}{c}{$67.1 \pm 2.3$} & \multicolumn{2}{c}{$61.5\pm0.5$} \\ \midrule

    \bottomrule
    \multirow{3}{*}{\textbf{PS}} &
           \multirow{3}{*}{Split} & \multicolumn{6}{c}{Testing Domains}        \\ \cmidrule(lr) {3-8}
        &&\multicolumn{2}{c}{NeuroSAT} & \multicolumn{2}{c}{GIN}  & \multicolumn{2}{c}{GCN}\\ \midrule
      
        \multirow{2}{*}{$\modularity$} & small & \multicolumn{2}{c}{$55.8 \pm4.0$} & \multicolumn{2}{c}{$55.1\pm0.1$} &\multicolumn{2}{c}{$57.3\pm0.6$}\\ & big & \multicolumn{2}{c}{$53.1\pm1.0$} & \multicolumn{2}{c}{$56.3\pm0.7$} &\multicolumn{2}{c}{$55.1\pm0.6$} \\  \midrule
 
    \bottomrule
    \multirow{3}{*}{\textbf{KCO}} &
           \multirow{3}{*}{Split} & \multicolumn{6}{c}{Testing Domains}        \\ \cmidrule(lr) {3-8}
        &&\multicolumn{2}{c}{NeuroSAT} & \multicolumn{2}{c}{GIN}  & \multicolumn{2}{c}{GCN}\\ \midrule
        \multirow{2}{*}{$\fractaldim$} & small & \multicolumn{2}{c}{$50.2\pm0.06$} & \multicolumn{2}{c}{$50.2\pm0.6$} & \multicolumn{2}{c}{$50.1\pm0.6$}  \\ & big & \multicolumn{2}{c}{$50.5\pm0.05$} & \multicolumn{2}{c}{$49.8\pm0.2$} & \multicolumn{2}{c}{$49.2\pm0.4$} \\ \midrule

        \multirow{2}{*}{$\treewidth$} & small & \multicolumn{2}{c}{$50.1\pm0.04$} & \multicolumn{2}{c}{$50.4\pm0.2$}  &\multicolumn{2}{c}{$49.5\pm0.8$}  \\ & big & \multicolumn{2}{c}{$64.5\pm1.4$} & \multicolumn{2}{c}{$50.1\pm0.1$} & \multicolumn{2}{c}{$49.0\pm0.4$}  \\ \midrule

        \multirow{2}{*}{$\modularity$} & small & \multicolumn{2}{c}{$50.2\pm0.04$} & \multicolumn{2}{c}{$48.0\pm0.5$} &\multicolumn{2}{c}{$50.3\pm0.3$}\\ & big & \multicolumn{2}{c}{$50.4\pm0.2$} & \multicolumn{2}{c}{$48.6\pm0.9$} &\multicolumn{2}{c}{$49.1\pm0.9$} \\  \midrule
        \multirow{2}{*}{$\Entropy$} & small & \multicolumn{2}{c}{$51.4\pm0.9$} & \multicolumn{2}{c}{$50.0\pm0.9$} & \multicolumn{2}{c}{$49.9\pm1.1$}\\
        & big & \multicolumn{2}{c}{$50.3\pm0.08$} & \multicolumn{2}{c}{$50.8\pm0.8$} & \multicolumn{2}{c}{$49.2\pm0.9$}  \\ 
    \bottomrule
    \end{tabular}}
        \vspace{-0.2cm}
        \label{tab:general3sat}
\end{table}

\begin{table}[t!]
        \centering
        \caption{Accuracy of NeuroSAT trained on SR(10-40) with different structural splits, task is satisfying assignment prediction.}
        \vspace{2pt}
        \resizebox{0.45\textwidth}{!}{\begin{tabular}{>{\centering\arraybackslash}p{0.46cm} >
        {\centering\arraybackslash}p{1cm} >
        {\centering\arraybackslash}p{0.65cm} >
        {\centering\arraybackslash}p{0.65cm} >{\centering\arraybackslash}p{0.65cm} >{\centering\arraybackslash}p{0.65cm} >{\centering\arraybackslash}p{0.65cm} >
        {\centering\arraybackslash}p{0.65cm} >
        {\centering\arraybackslash}p{0.65cm} >
        {\centering\arraybackslash}p{0.65cm} >
        {\centering\arraybackslash}p{0.65cm} >
        {\centering\arraybackslash}p{0.65cm} >
        {\centering\arraybackslash}p{0.65cm}}
        \toprule
           \multirow{3}{*}{Metric} &
           \multirow{3}{*}{Split} & \multicolumn{11}{c}{Testing Domains}        \\ \cmidrule(lr) {3-13}
        &&SR & R3 & KCL & KD  & KV & KCO & AM & CA & PS & G2 & IN \\ \midrule
        \multirow{2}{*}{$\scalefreev$} & small & 94.40 & 91.72 & 48.76 & 66.03& 74.89 & 44.70 &  46.98 & 90.28 & 94.44 & 95.83 & 50.00 \\  & big & 93.05 & 91.40 &51.60 & 53.21 & 53.64 & 56.14 & 58.37& 70.51 & 91.11 & 83.33 & 50.00\\ \midrule
        \multirow{2}{*}{$\scalefreec$} & small& 95.41 & 92.99 & 51.85  & 59.27 & 68.10 & 55.97 & 43.85 & 92.36 & 96.04 & 87.50 & 58.33\\ & big & 95.31 & 92.82 & 48.83 & 56.14 & 54.97& 51.75& 45.79 & 93.31 & 96.06 & 91.67 & 75.00\\  \midrule
        \multirow{2}{*}{$\modularity$} & small & 93.76 & 91.87  & 56.30  & 71.76 & 51.9 & 55.89 & 49.75 & 89.08 & 95.75 & 0.79 & 50.00\\ & big & 94.81 & 92.74 & 51.61 & 61.06 & 60.80 & 61.64 & 46.57 & 89.99 & 93.00 & 83.33 & 58.33\\  \midrule
    \bottomrule
    \end{tabular}}
        \vspace{-0.2cm}
        \label{tab:general111}
\end{table}

\end{appendix}
\end{document}